\documentclass[12pt]{article}
\usepackage{amsmath}
\usepackage{times}
\usepackage{graphicx}
\usepackage{color}
\usepackage{multirow}
\usepackage[authoryear]{natbib}
\usepackage{lineno}

\usepackage{rotating}
\usepackage{bbm}
\usepackage{latexsym}

\usepackage{algorithm} 
\usepackage{algorithmic}
\usepackage{wrapfig}

\usepackage{amsthm}

\usepackage{amssymb,amsfonts}

\usepackage{multirow}
\usepackage{makecell}
\usepackage{newtxtext,newtxmath}
\usepackage{graphicx}
\usepackage{algorithm}
\usepackage[algo2e]{algorithm2e}
\usepackage{algorithmic}
\usepackage{marvosym}
\usepackage{epstopdf}

\usepackage{amssymb}

\usepackage{color}
\usepackage{colortbl}
\usepackage{hyperref}
\usepackage{subfigure}
\usepackage{enumerate}

\usepackage[inline, shortlabels]{enumitem}
\setlist{itemjoin ={,\enspace},itemjoin* = { and\enspace}}

\usepackage{booktabs}       

\usepackage{mathtools}
\usepackage{pifont}
\newcommand{\cmark}{\ding{51}}%
\newcommand{\xmark}{\ding{55}}%
\usepackage{extarrows}
\usepackage{color}
\usepackage{colortbl}
\definecolor{mygray}{gray}{.8}
\definecolor{myblue}{rgb}{0.61, 0.87, 1.0}

\usepackage{tikz}
\usepackage{etoolbox}

\newcommand{\circled}[2][]{\tikz[baseline=(char.base)]
	{\node[shape = circle, draw, inner sep = 1pt]
		(char) {\phantom{\ifblank{#1}{#2}{#1}}};%
		\node at (char.center) {\makebox[0pt][c]{#2}};}}
\robustify{\circled}

\usepackage{soul}
\soulregister\cite7 
\soulregister\citep7 
\soulregister\citet7 
\soulregister\ref7 
\soulregister\pageref7 

\newcommand{\captionfonts}{\normalsize}

\makeatletter  
\long\def\@makecaption#1#2{%
	\vskip\abovecaptionskip
	\sbox\@tempboxa{{\captionfonts #1: #2}}%
	\ifdim \wd\@tempboxa >\hsize
	{\captionfonts #1: #2\par}
	\else
	\hbox to\hsize{\hfil\box\@tempboxa\hfil}%
	\fi
	\vskip\belowcaptionskip}
\makeatother   

\begin{document}
	
	\hspace{13.9cm}1
	
	\ \vspace{20mm}\\
	\begin{center}
		{\LARGE Multi-view Alignment {and Generation in CCA} \\ via Consistent Latent Encoding}
	\end{center}
	\ \\
	{\bf \large Yaxin Shi$^{\displaystyle 1}$, Yuangang Pan$^{\displaystyle 1}$ and Donna Xu$^{\displaystyle 1}$ and Ivor W. Tsang$^{\displaystyle 1}$}\\
	{$^{\displaystyle 1}$ Centre for Artificial Intelligence (CAI), University of Technology Sydney, Australia.}\\
	%
	
	
	\thispagestyle{empty}
	\markboth{}{NC instructions}
	\ \vspace{-0mm}\\
	%
	\begin{center} {\bf Abstract} \end{center}
	{Multi-view alignment, achieving one-to-one correspondence of multi-view inputs, is critical in many real-world multi-view applications, {especially for {cross-view data analysis problems.}} Recently, an increasing number of works study this alignment problem with Canonical Correlation Analysis~(CCA). However, existing CCA models are prone to misalign the multiple views due to either the neglect of uncertainty or the inconsistent encoding of the multiple views. To tackle these two issues, this paper studies multi-view alignment from the Bayesian perspective. {Delving into the impairments of inconsistent encodings, we propose to recover correspondence of the multi-view inputs by matching the marginalization of the joint distribution of multi-view random variables under different forms of factorization. To realize our design, we present Adversarial CCA (ACCA) which achieves consistent latent encodings {by matching the marginalized latent encodings through the adversarial training paradigm.}} Our analysis based on conditional mutual information reveals that ACCA is flexible for handling implicit distributions. Extensive experiments on correlation analysis and cross-view generation under noisy input settings demonstrate the \mbox{superiority of our model}.
%
	
	\section{Introduction}\label{sec:introduction}
	Multi-view learning is the subfield of machine learning that considers learning from data with multiple feature sets. This paradigm has attracted increasing attention due to the emerging multi-view data that have facilitated various real-world applications, such as video surveillance~\citep{wang2013intelligent}, information retrieval~\citep{elkahky2015multi} and recommender systems~\citep{elkahky2015multi}. {In these applications, it is critical to achieve \textbf{\emph{instance-level multi-view alignment}}, such that the multiple data streams achieve great one-to-one correspondence~\citep{li2018survey}}. For example, considering traditional multi-view learning tasks, e.g. multi-view classification~\citep{qi2016volumetric}, multi-view clustering~\citep{chaudhuri2009multi} on face images in video surveillance, the input data corresponds to face images taken from different angles. In these case, input feature sets with low one-to-one correspondence degrade the alignment of the multiple views, thus severely affect the performance of the desired tasks. Furthermore, multi-view alignment plays an even more critical role in \textbf{\emph{cross-view data analysis}} \citep{jia2016cross} {problems}, namely, to analyse one view of the data given the input from the other view. For example, cross-view retrieval~\citep{elkahky2015multi} aims to search for the corresponding object in the target view by given the quay in the other view; cross-view generation~\citep{regmi2018cross} seeks to generate target objects given the cross-view inputs. Both of them are promising real-world application in which alignment of the incorporated views is critical for the performance. 
	
	\emph{Canonical Correlation Analysis (CCA)}~\citep{hotelling1936relations} provides a primary tool to study instance-level multi-view alignment under subspace learning mechanism~\citep{xu2013survey}. {In this setting, the instances of two views, $X$ and $Y$, are assumed to be generated from a common latent subspace $\mathcal{Z}$, the alignment problem is to find two mapping functions, namely $F(X)$ and $G(Y)$, such that the embeddings of corresponding input pairs are close to each other regarding the linear correlation. The instance $(x_{i}, y_{i})$ are in exact correspondence if and only if $F(x_{i}) = G(y_{i})$~\citep{ma2011manifold}.} However, existing CCA models are prone to misalignment, due to either the neglect of uncertainty or the inconsistent encoding of the multiple views.
	
	Following the principle of classic CCA, \textbf{\emph{vanilla CCA models}} study multi-view alignment with deterministic mapping functions~\citep{oh2018modeling}. Such CCA models are opting to misalign the multiple views since \textbf{\emph{uncertainty}} is not considered. To be specific, the classic CCA obtains the shared latent space by {maximumly correlating} the deterministic point embeddings, achieved with a linear mapping of the two views. 
	Some works, such as \emph{Kernel CCA (KCCA)}~\citep{lai2000kernel} and \emph{Deep CCA (DCCA)}~\citep{andrew2013deep} and ~\emph{Multi-View AutoEncoder (MVAE)}~\citep{ngiam2011multimodal}, extend the classic CCA with nonlinear mapping or through cross-view reconstruction, to exploit nonlinear correlation for the alignment. The mapping functions $F(\cdot)$ and $G(\cdot)$ are nonlinear in these models. As depicted in Fig.~\ref{fig:motivation}(a), these methods all exploit the subspace $Z$ with deterministic point embeddings, namely \begin{small}$z_{x} = F(x)$
	\end{small} and \begin{small} $z_{y} = G(y)$\end{small} are points in \begin{small}$\mathbb{R}^{{d}}$\end{small}. {Without an inference mechanism to evaluate the quality of obtained latent codes, the mapping function obtained in those models is susceptible to \emph{noisy inputs~\citep{kendall2017uncertainties}}, which can consequently {\emph{result in misalignment of the multiple views}}. For example, for observation ``{\small \circled[2]{1}}'' in Fig.~\ref{fig:motivation}(a), inputs in the two views are obviously projected faraway in the embedding space - they are projected into different clusters, 5 and 2 respectively, while they are suppose to be close to each other around the ground truth cluster 7.} Moreover, without prior regularization on the shared subspace, these models can not allow easy latent interpolations, since their latent spaces are \emph{discontinuous}. In such cases, the training samples are encoded into non-overlapping zones chaotically scattered across space, with {``holes''} between the zones where the model has never been trained~\citep{tolstikhin2017wasserstein}. Therefore, these models {\emph{can not facilitate the cross-view generation task}} since the generation results are quite likely to be unrealistic.
	\begin{figure}[t]
		\centering
		\includegraphics[width=0.99\textwidth]{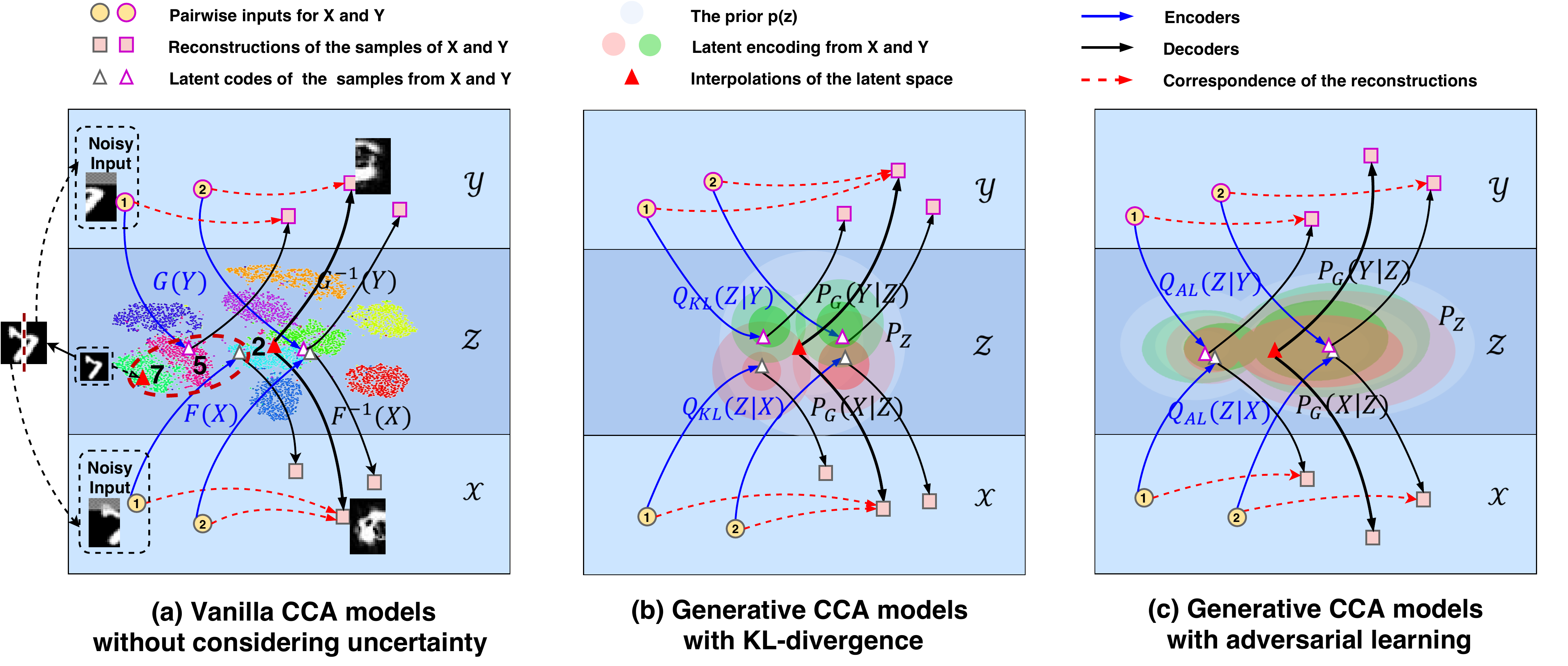}
		\vspace{-2.5mm}\caption{The motivation of Adversarial CCA. \textbf{(a)} Vanilla CCA models misalign the multiple views with discontinuous latent space and unrealistic generated data. \textbf{(b)} The latent encodings matched with KL-divergence are inconsistent, leading to misalignment of the multiple views. \textbf{(c)} Adversarial learning facilitates consistent encodings for the multiple views by matching marginalized latent encodings with flexible priors.
		}
		\vspace{-4mm}
		\label{fig:motivation}
	\end{figure} 
	
	\textbf{\emph{Generative CCA models}}, such as~\emph{probabilistic CCA (PCCA)}~\citep{bach2005probabilistic},~\emph{Variational CCA (VCCA)}~\citep{wang2016deep} and~\emph{Multi-Channel Variational Autoencoder (MCVAE)}~\citep{antelmi2019sparse}, overcome the aforementioned issue with probability. However, they suffer from misalignment due to the impairments of \textbf{\emph{inconsistent encodings}}. Specifically, these models adopt the Kullback-Leibler divergence~(\emph{KL-divergence}) between the encodings of individual input example, i.e. $Q(Z| X = x)$ and $Q(Z| Y = y)$ and the prior $P_{0}(Z)$, as the criterion to match the latent encodings of different views. However, such constraint can simply force the matching of the encodings of individual input to the common prior~\citep{tolstikhin2017wasserstein}. Even if the constraint is satisfied, the encodings of the data samples from both the two views can be intersected. In this way, the correspondence between the latent codes of paired inputs is violated. {Such inconsistent latent encodings would cause one-to-many correspondence between the instances of the incorporated views, indicating the multiple views are misaligned.} As depicted in Fig.~\ref{fig:motivation}(b), although all these latent encodings match the prior, the encodings of the instances from both the two views are intersected in the common latent space. This arouses confusion on the correspondence between the instances in the two views, e.g.``{\small \circled[2]{1}}'' and ``{\small \circled[2]{2}}'' {both exhibit \textbf{one-to-many}} correspondence. Such inconsistency not only weakens the alignment of the two spaces but also influences the quality of data reconstruction. Moreover, to achieve a tractable solution for the inference, these models restrict the latent space with simple Gaussian prior, i.e. $p_{0}(z)\sim \mathcal{N}(0, I_{d})$, so that the constraint can be computed analytically. However, such prior is not expressive enough to capture the true posterior distributions~\citep{mescheder2017adversarial}. Therefore, the latent space may not be expressive enough to preserve the instance-level correspondence of the data samples. These impairments lead to an {\emph{inferior alignment of the multiple views and thus also {degrade} the models' performance in cross-view generation tasks.}}
	
	To tackle the aforementioned issues, in this paper, we study the instance-level multi-view alignment from a Bayesian perspective. With an in-depth analysis of existing CCA models with respect to latent distribution matching, we figure out the impairments of inconsistent encodings in the existing CCA models. We then propose to recover consistency of multiple views {and thereby boost the cross-view generation performance}, by matching the  \textbf{\emph{marginalization}} of the joint distribution of multi-view random variables under different forms of factorization, i.e. Eq.~\eqref{eq:ACCA_three_appro}. To realize our marginalization design, we present Adversarial CCA (ACCA) which achieves consistent latent encoding of the multiple views by matching the marginalized posteriors to flexible prior distributions { through the adversarial training paradigm}. Analysing the conditional independent assumption in CCA with conditional mutual information (CMI), we reveal that, compared with existing CCA methods, our ACCA is flexible for handling implicit distributions. The contributions of this work can be summarized as follows:
	\begin{table}[t]
		\begin{center}
			\renewcommand{\arraystretch}{1.2}
			\caption{Comparison of different CCA methods for multi-view alignment.}
			\vspace{1mm}
			\label{tab:CCAvariants}
			\setlength{\tabcolsep}{1.2mm}{
				\scalebox{0.72}{
					\begin{tabular}{c|c|c|c|>{\columncolor{mygray}}ccc}
						\toprule				
						\multirow{3}{*}{\textbf{Category}}&\multirow{3}{*}{\textbf{Methods}} & \multirow{3}{*}{\begin{tabular}[c]{@{}c@{}}\textbf{Nonlinear}\\ \textbf{ mapping}\end{tabular}} & \multirow{3}{*}{\textbf{Criterion}} & \multicolumn{3}{c}{\textbf{ Evaluation}} \\ \cline{5-7}
						\multicolumn{1}{c|}{} & & & &{\begin{tabular}[c]{@{}c@{}}\textbf{Consistent}\\  \textbf{encoding}\end{tabular}} & {\begin{tabular}[c]{@{}c@{}}\textbf{Avoids Gaussian}\\  \textbf{restriction on $p(\mathbf{z})$}\end{tabular}} & {\begin{tabular}[c]{@{}c@{}}\textbf{Implicit}\\  \textbf{posteriors $p(\mathbf{z}|\mathbf{x},\mathbf{y})$}\end{tabular}}  \\ \hline
						\multirow{5}{*}{\begin{tabular}[c]{@{}c@{}}\textbf{Vanilla}\\ \textbf{CCA models}\end{tabular}}
						&CCA    & \xmark &Linear correlation& \xmark & \xmark& \textbf{-}\\
						&KCCA   & \cmark &Linear correlation& \xmark & \xmark& \textbf{-}\\
						&DCCA   & \cmark &Linear correlation& \xmark & \xmark& \textbf{-}\\
						&DCCAE  & \cmark &Linear correlation& \xmark & \xmark& \textbf{-}\\
						&MVAE   & \cmark &\textbf{-}& \textbf{-} & \textbf{-} &\textbf{-} \\ \hline
						\multirow{5}{*}{\begin{tabular}[c]{@{}c@{}}\textbf{Generative}\\ \textbf{CCA models}\end{tabular}}
						&PCCA       & \xmark &KL-divergence& \xmark & \xmark & \xmark\\
						&VCCA       & \cmark &KL-divergence& \xmark & \xmark & \xmark\\
						&Bi-VCCA    & \cmark &KL-divergence& \xmark & \xmark & \xmark\\
						&MCVAE      & \cmark &KL-divergence& \xmark & \xmark & \xmark\\
						&\textbf{ACCA (ours)}& \cmark &Adversarial learning  & \cmark & \cmark & \cmark\\
						\bottomrule
			\end{tabular}}}
		\end{center}
		\vspace{-5mm}
	\end{table}
	\begin{itemize}
		\item [1.] {We provide a systematic study on CCA-based instance-level multi-view alignment. We figure out the impairments of inconsistent encodings in the existing CCA models and propose to study multi-view alignment based on the marginalization principle of Bayesian inference, to recover consistency of multiple views.}
		\item [2.] {We design adversarial CCA (ACCA) which achieves consistent latent encoding of the multiple views and is flexible for handling implicit distributions. To the best of our knowledge, we are the first to elaborate the superiority of adversarial learning in multi-view alignment scenario.}
		\item [3.] We analyse the connection of ACCA and existing CCA models based on CMI and reveals the superiority of ACCA benefited from the consistent latent encoding. Our CMI-based analysis and the consistent latent encoding can provide insights for a flexible design of other CCA models for multi-view alignment.
	\end{itemize}
	The rest of this paper is organized as follows. In Section~\ref{sec:related_work}, we review the existing CCA models regarding latent distribution matching. In Section~\ref{sec:ACCA}, we elaborate our design to study multi-view alignment through marginalization and present our design of Adversarial CCA (ACCA). In Section~\ref{sec:discussion}, we discuss the advantages of our model by comparing existing models based on CMI. In Section~\ref{sec:experiments}, we demonstrate the superior alignment performance of ACCA with model verification and various real-world applications. Section~\ref{sec:conclusions} concludes the paper and envisions future work.
	
	\section{Deficiencies of existing CCA models }\label{sec:related_work} 
	In this section, we review the multi-view alignment achieved with existing CCA models in terms of latent distribution matching.
	
	\subsection{Vanilla CCA models and the neglect of uncertainty}\label{sec:vanillaCCA}
	Vanilla CCA models are prone to misalignment since data uncertainty is not considered.
	
	\vspace{0.5mm}
	
	\emph{Canonical Correlation Analysis (CCA)} ~\citep{hotelling1936relations} is a powerful statistical tool for multi-view data analysis. Let $\{x^{(i)}, y^{(i)}\}_{i=1}^{N}$ denote the collection of $N$ i.i.d. samples with pairwise correspondence in multi-view scenario (In the following, we use $(x,y)$ to denote any one instance in this set, for simplicity). The classic CCA aims to find linear projections for the two views, ($W_{x}^{'}X, W_{y}^{'}Y$), such that the correlation between the projections are mutually maximized, namely~\mbox{\begin{small}$\max \; corr\{W_{x}^{'}{X}, {W}_{y}^{'}{Y}\}=\frac{{W}_{x}^{'}\mathbf{\Sigma}_{xy}{W}_{y}}{\sqrt{{W}_{x}^{'}\mathbf{\Sigma}_{xx}{W}_{x}{{W}_{y}^{'}\mathbf{\Sigma}_{yy}{W}_{y}}}}$\end{small}}, where ${\mathbf{\Sigma}_{xx}}$ and  ${\mathbf{\Sigma}}_{yy}$ are the covariance of $X$ and $Y$; ${\mathbf{\Sigma}}_{xy}$ denotes the cross-covariance. With linear projections, the classic CCA simply exploits linear correlation among the multiple views to achieve alignment. It is often insufficient to analyse complex real-world data that exhibits higher-order correlations~\citep{suzuki2010sufficient}.
	
	Various CCA models are proposed to exploit nonlinear correlation for multi-view alignment with deterministic nonlinear mappings. \emph{Kernel CCA (KCCA)} and \emph{Deep CCA (DCCA)} exploit nonlinear correlation by extending CCA with nonlinear mapping implement with kernel methods and Deep Neural Networks (DNNs), respectively. Some other works, e.g.~\emph{deep canonically correlated autoencoders (DCCAE)}~\citep{wang2015deep}, extend nonlinear CCA with self-reconstruction for each view. However, since there is a trade-off between canonical correlation of the learned bottleneck representations and the reconstruction, the cross-view relationship captured in the common subspace is often inferior to that of DCCA~\citep{wang2016deep}. \emph{Multi-View AutoEncoder} (MVAE) aims to establish strong connection between the views through cross-view reconstruction. Without adopting specific alignment criterion, its objective is given as 
	\begin{equation}
	\min \limits_{F, G} \frac{1}{N} \sum_{\{x,y\}} {\| x - F^{-1}(F(x))\|}^{2} + {\| x - F^{-1}(G(y))\|}^{2} + {\| y - G^{-1}(G(y))\|}^{2} + {\| y -G^{-1}(F(x))\|}^{2}, \nonumber \label{eq:MVAE}
	\end{equation}
	where $F(.)$ and $G(.)$ represent nonlinear mapping of $X$ and $Y$ respectively. $F^{-1}(.)$ and $G^{-1}(.)$ denote the corresponding decoders for the view reconstructions.  
	
	\vspace{0.5mm}
	
	\textbf{Necessities of modelling uncertainty:} {Without the inference mechanism that can evaluate the quality of obtained embeddings, these methods are vulnerable to misalign the multiple views when given noisy inputs~\citep{tolstikhin2017wasserstein}. As depicted in Fig.~\ref{fig:motivation}(a), for noisy halved images of digit ``7'', the two views are misaligned in the latent space, since their embeddings scatter faraway and are even chaotically embedded into the different clusters of ``2'' and ``5'', respectively.}  Moreover, these models can not well facilitate cross-view generation tasks, since the obtained subspace is discontinuous under such deterministic mappings. Consequently, interpolations of the latent space would lead to unrealistic generation results.
	
	\subsection{{Generative CCA models and inconsistent latent encodings}}\label{sec:generativeCCA}
	
	Generative CCA models overcome the uncertainty issue by modeling probability. However, they still suffer from misalignment due to the impairments of inconsistent encodings, caused by the limitation of the \emph{KL-divergence} alignment criterion.
	
	\vspace{0.5mm}
	
	Let the two input views correspond to random variables $X$ and $Y$, each of them are distributed according to an unknown generative process with density $p(x)$ and $p(y)$ from which we have observations $\{x^{(i)}, y^{(i)}\}_{i=1}^{N}$. \emph{Probabilistic CCA (PCCA)}~\citep{bach2005probabilistic}, as generative version of the classic CCA, aligns the multi-view data by maximizing the correlation between the linearly projected views in a common latent space with Gaussian prior, namely ${z}\sim\mathcal{N}({0},{I}_{d}),~
	{x}|{z}\sim \mathcal{N}({W}_{x}{z}+{\mu}_{x},{\Phi_{x}}),~
	{y}|{z}\sim\mathcal{N}({W}_{y}{z}+{\mu}_{y},{\Phi_{y}}),$
	where $d$ denotes the dimension of the projected space. The KL-divergence is tractable in this case, since the conjugacy of the prior and the likelihood in PCCA leads to two favorable conditions. \emph{1).} the conditional distribution $p({x},{y}|{z})$ can be modeled with the joint covariance matrix, with which the conditional independent constraint (Eq.~\eqref{eq:conditional_independent}) for CCA can be easily imposed~\citep{drton2008lectures} .
	\begin{equation}
	p{({x},{y}|{z})} = p{({x}|{z})}p{({y}|{z})}\label{eq:conditional_independent}
	\end{equation}
	\emph{2).} the posterior, i.e. \begin{small}$p{({z}|{x},{y})} = \frac{p{({x},{y}|{z})}p({{z}})}{p{({x},{y})}}$\end{small} can be calculated analytically~\citep{tipping1999probabilistic}. 
	
	To exploit nonlinear correlation for alignment, some works extend PCCA with nonlinear mapping. Inspired by variational inference, {Wang et al. proposed} two generative CCA variants~\citep{wang2016deep}, \emph{Variational CCA} and \emph{Bi-VCCA}. Both methods minimize a reconstruction cost together with the  KL-divergence to regularize the alignment. \emph{Variational CCA (VCCA)} penalizes the discrepancy between a single view encoding and the prior, i.e. $D_{KL}(Q(Z|X=x) \parallel P_{0}(Z))$, based on a preference for one of the two views. The two views are not well aligned since the information in the other view is not exploited. It also cannot handle the cross-view generation task due to this missing encoding. \emph{Bi-VCCA} overcomes the limitation by a heuristic combination of the KL-divergence term obtained with both the two encodings, $Q(Z| X = x)$ and $Q(Z| Y = y)$, with $\lambda$ to control the trade-off. To achieve tractable solution for the inference, the latent space is restricted to be Gaussian distributed, i.e. $P_{0}(Z) \sim \mathcal{N}(\mu,\Sigma)$, so that the KL-divergence can be computed analytically. Its objective is given as
	\begin{eqnarray}
	\lefteqn{\min \limits_{\theta,\phi} \frac{1}{N} \sum_{\{x,y\}} {\big[}\lambda  [-\mathbb{E}_{q_{\phi}(z|x)}[\log {p_{\theta}({x}|{z})}+\log {p_{\theta}({y}|{z})}] 
		+{D_{KL}(q_{\phi}({z}|{x})\parallel p_{0}({z}))}]} \\ \nonumber
	&& \qquad \qquad+ (1-\lambda)  [-\mathbb{E}_{q_{\phi}(z|y)}[\log {p_{\theta}({x}|{z})}+\log {p_{\theta}({y}|{z})}] + {D_{KL}(q_{\phi}({z}|{y})\parallel p_{0}({z}))}]{\big]}, \nonumber \label{eq:BIVCCA}
	\end{eqnarray}
	where $\theta$ is the generative model parameters, $\phi$ denotes the variational parameters. 
	The prototype proposed in ~\cite{antelmi2019sparse}, namely \emph{Multi-Channel Variational Autoencoder (MCVAE)}, aims to constraint the expectation of KL-divergence between the encoding of each view and the target posterior distribution, i.e. $ Q(Z|X = x)$, $Q(Z|X = y)$ and $Q(Z|X = x, Y = y)$ for all the data samples, as the criteria for the alignment. However, with an explicit \textbf{\emph{conditionally independent assumption}} (Eq.~\eqref{eq:conditional_independent}), MCVAE achieves the same objective as Bi-VCCA.
	
	\textbf{Impairments of inconsistent latent encodings:}
Since there exists an encoding and decoding mechanism for each of the views in generative CCA models, the instance-level alignment of the views can be verified by cross-view generation. Specifically, if the two views are well aligned, the encoding from one view can then recover the corresponding data in the other view. In such circumstances, we define the encoding of the two views to be \textbf{\emph{``consistent''}}. Therefore, the consistency of the multi-view encodings is a necessary condition for multi-view alignment in generative CCA models.

	However, the aforementioned methods would misalign the multiple views due to the inconsistent latent encodings caused by the inferior alignment criterion, i.e. $D_{KL}(Q(Z|X) \parallel Q(Z|Y))$. First, this criterion can only match the encodings of individual data samples, while causing inconsistent encoding of the views. As depicted in Fig.~\ref{fig:motivation}(b), in the multi-view learning scenario, it simply forces the encoding from each view, i.e. $Q(Z| X = x)$ and $Q(Z| Y = y)$, of all the different input examples to individually match the common prior $P_{0}(Z)$. In this way, the latent encodings from both the two views are intersected in the common latent space. {Such intersection disorganizes the consistency of the encodings in the latent space, and thus reduce the instance-level alignment of the two input views. This misalignment also influences the quality of data reconstruction or generation. Both the two deficiencies are crucial for cross-view generation tasks.} In addition, to compute the KL-divergence analytically, all these methods require the incorporated distributions, i.e. the prior $P_{0}(Z)$, the posteriors of each view $Q(Z|X)$ and $Q(Z|Y)$, to be simple. However, such restriction can lead to inferior inference models that are not expressive enough to capture the true posterior distribution~\citep{mescheder2017adversarial}. Inexpressiveness of the latent space further limits the models' ability to preserve the instance-level correspondence of the data samples.
	\begin{figure}[t]
		\centering
		\includegraphics[width=0.9\textwidth]{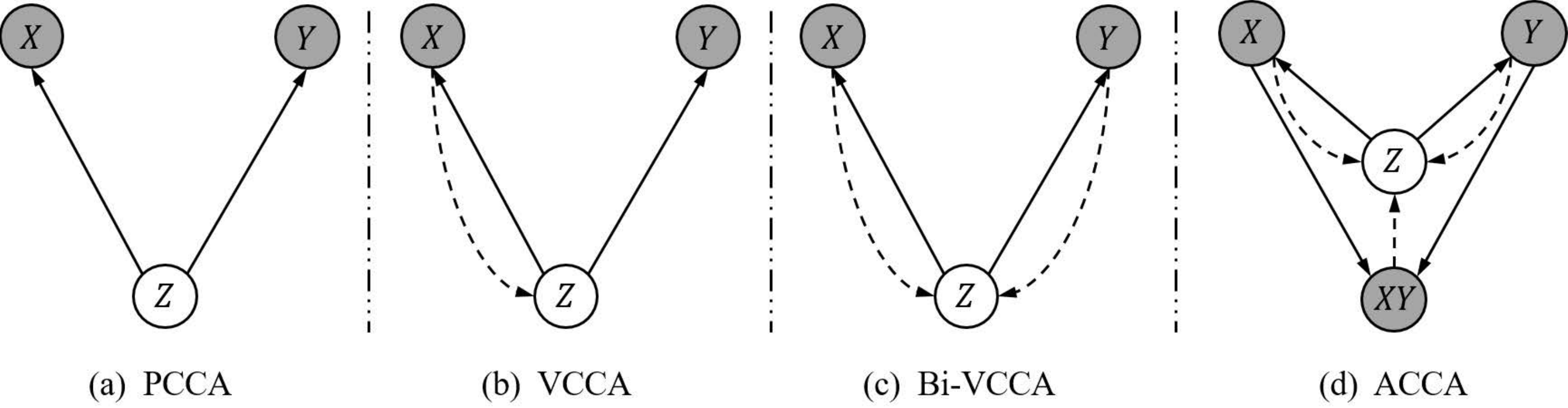}
		\caption{Graphical diagrams for generative nonlinear CCA variants. The solid lines in each diagram denote the generative models $p_{\theta}({z})p_{\theta}({*}|{z})$. The dashed lines denote the approximation $q_{\phi}({z}|{*})$ to the intractable posterior $p_{{\theta}}({z}|{*})$. The $*$ indicates $x$ or $y$. }\label{fig:diagrams}
	\end{figure}
	\section{Multi-view alignment via consistent latent encoding}\label{sec:ACCA}
	In this section, we study multi-view alignment from a Bayesian perspective. First, we elaborate the design to achieve consistency of the multiple views through marginalization in Section~\ref{sec: marginalization}. We then present our design of adversarial CCA in Section~\ref{sec:consistent_latent_encoding}.
	
	\subsection{{Multi-view alignment through marginalization}}\label{sec: marginalization}
	To sum up, the KL-divergence criterion adopted in existing CCA models causes impairments of the inconsistent encodings in two aspects:
	\begin{itemize}
		\item [1.] Primarily, it causes inconsistent latent encoding of the two views, since it simply matches the encodings of individual data samples; 
		\item [2.] It further restricts the expressiveness of the latent space regarding the instance-level correspondence, since it can only incorporate simple priors directly. 
	\end{itemize}
	To exploit a better criterion that benefits the alignment, i.e. instance-level consistency, of the multiple views, we study multi-view alignment from a Bayesian perspective.
	
	From the Bayesian perspective, the primary reason for the inconsistent encoding is that their KL-divergence criterion measures the disagreement of the posterior distributions $q(z|x)$ and $q(z|y)$ \textbf{\emph{without considering the condition variable.}} That is, it simply matches the encodings of individual data in each view to the prior $p_{0}(z)$ via a heuristic combination of the KL-divergence between each encoding and the prior, namely
	\begin{equation}
	\lambda  D_{KL}(q(z|x)\parallel p_{0}(z)) + (1-\lambda) D_{KL}(q(z|y)\parallel p_{0}(z)).\label{eq:metric_heuristic_combination}
	\end{equation}
	{Without considering the condition variables $X$ and $Y$, the encodings of instances from both the two views can be disorganized overlapped. This degrades the one-to-one correspondence of the multi-view data in corresponding models.}
	
	{Based on the marginalization principle of Bayesian inference~\citep{tipping2003bayesian,jaynes1978marginalization}, we propose to facilitate consistent latent encoding by \textbf{\emph{simultaneously}} matching the multi-view encodings whose condition variables are all integrated out. Specifically, we first eliminate the misalignment induced by the intersection of the individual sample encodings by marginalizing the encodings from multiple views and then constrained the marginalized encodings to overlap with the prior $p_{0}(z)$ simultaneously.}
	
	{First,  within the CCA-based multi-view learning scenario, the joint distribution of multi-view random variables can be factorized into three different forms, i.e. $q({x}, {z})= q({z}|{x})p({x}), q({y}, {z})= q({z}|{y})p({y}),~q({{x}, {y}},{z})= q({z}|{x},{y}) p({x},{y})$. Marginalization of these joint distributions on $z$ results in three marginalized posterior distributions, namely } 
	\begin{equation}
	q_{{x}}({z})=\int q({z}|{x})p({x})\,d{x},\;q_{{y}}({z})=\int q({z}|{y})\,p({y})\;d{y},~q_{{x}{y}}({z})=\iint q({z}|{x},{y}) p({x},{y})d{x}d{y}.\label{eq:marginalization}
	\end{equation}
	{Then, we propose to match these three marginalized encodings simultaneously, to provide consistent latent encodings that benefit the multi-view alignment. Since it is non-trivial to annotate a distribution measurement among the prior $p_{0}(z)$ and other surrogate distributions marginalized by different views, we represent this idea as}
	\begin{equation}\label{eq:ACCA_three_appro}
	q_{{x}}({z}) \approx 
	q_{{y}}({z}) \approx
	q_{{x}{y}}({z}) \approx p_{0}({z}), 
	\end{equation} 
	
	{Compared with the KL-divergence that harshly matches the conditional distribution of each sample to the prior, our proposed constraint matches the marginal distributions, i.e. $\int q(z|x)p(x)d{x}\approx p(z)$. Since we take the input of the conditional variables into consideration, this constraint is tolerant to the flexibility of the input data. This property also makes it praisable for matching multi-view encodings, i.e. $\int q(z|x)p(x)d{x} \approx \int q(z|y)p(y)d{y} \approx \int q(z|x, y)p(x, y)d{x}d{y} \approx p_{0}(z)$. The multi-view alignment can be further improved via expanding the expressiveness of the latent space by \textbf{\emph{incorporating more complex prior distributions}}~\citep{DBLP:conf/icml/MathieuRST19}.}  
	
	\subsection{{Adversarial CCA with consistent latent encoding}}\label{sec:consistent_latent_encoding}
	To realize our design, we design Adversarial CCA (ACCA) which provides consistent latent encoding by matching the marginalized latent encodings to flexible priors {through the adversarial training paradigm}. We adopt two schemes to facilitate consistent latent encodings in ACCA.
	
	\vspace{1mm}
	
	\noindent \textbf{Encoding with holistic information}: To provide different factorization forms for the joint distribution of multi-view data, we provide holistic information for the latent encodings, i.e. $q({z}|{x},{y})$, $q({z}|{x})$ and $q({z}|{y})$, in ACCA.
	
	\vspace{0.5 mm}
	
	Besides the two principle encodings, i.e. $q({z}|{x})$ and $q({z}|{y})$, that support the cross-view analysis, we further explicitly model $q({z}|{x},{y})$ by encoding an auxiliary view $XY$ that contains all the information of the two views. With the encoding from this auxiliary view, the latent space is more expressive for the correspondence of the multiple views.
	
	\vspace{0.5mm}
	
	\noindent {\textbf{Matching marginalized encodings}: We match the marginalization of these holistic encodings simultaneously with the adversarial learning technique.
		
	\vspace{0.5 mm}
	{The adversarial learning technique {minimizes} the JS-divergence between two distributions through binary classification on the samples of the two distributions directly~\citep{goodfellow2014generative}}. Consequently, any two distributions can be matched by given their samples. {We adopt adversarial learning as the criterion to match the marginalization of all three encodings to an {arbitrary fixed prior} $p_{0}({z})$ in ACCA. To be specific, we apply an adversarial distribution matching scheme on the common latent space. Within this scheme, each encoder acts as a generator that defines a marginalized posterior over~${z}$~\citep{makhzani2015adversarial} in Eq.~\eqref{eq:marginalization}. The obtained latent codes of individual data instances are samples of the corresponding marginalized posteriors, $q_{*}(z)$. The three marginalized posteriors constraint to be matched by simultaneously matching the same prior $p_{0}({z})$, namely Eq.~\eqref{eq:ACCA_three_appro}, with a \textbf{\emph{shared discriminator}}~\citep{hoang2018mgan}.} We presents the formulation of the proposed constraints in the following subsection.  
		
	Consequently, our ACCA realizes the proposed marginalization design by adversarially matching the marginalized posteriors with a common and flexible prior distribution. As listed in Table~\ref{tab:CCAvariants}, our ACCA excels existing generative CCA models in three aspects. 
		\begin{itemize}	
			\item[1.] We recover the consistency of multiple views by matching the \textbf{\emph{marginalization}} of holistic encodings. This inherence contributes to the consistent latent encoding of the multiple views that benefits the multi-view alignment.
			\item[2.] It avoids the Gaussian distribution restriction on $p(z)$. Instead of computing the criterion analytically, adversarial learning provides an efficient estimation of the JS-divergence between the encodings~\citep{goodfellow2014generative}. This benefits ACCA to handle expressive latent space with flexible prior distributions.
			\item[3.] It does not require explicit distribution assumptions on the posterior $p(z|x,y)$. The adversarial learning scheme matches the incorporated distributions implicitly. Thus it can benefit the model to omit the sampling operation required in other generative CCA models, e.g. VCCA and MCVAE.
		\end{itemize}
		
		The graphical diagram of ACCA is presented in Fig.~\ref{fig:diagrams}(d). Note that, \emph{the three encodings are all essential in ACCA.} First, the encodings of the principle views, i.e. $q({z}|{x})$ and $q({z}|{y})$, are essential to facilitate cross-view analysis with generative CCA methods. Second, the encoding of the auxiliary view, $q({z}|{x, y})$, contributes to a latent space that better encodes the correspondence of the multiple views and thus benefits the multi-view alignment achieved in ACCA. Indeed, one can achieve expressive representations for the multi-view data with only the auxiliary encoding. However, this is not the focus of our work. We further emphasise the significance of the auxiliary view and the superiority achieved with the adversarial learning in Section~\ref{sec: ACCA_superiority}.
		
		\subsubsection{Formulation}\label{sec:ACCA_formulation}
		
		{Based on the aforementioned design, the objective of our ACCA consists of two components: \begin{enumerate*}
			\item[1).] The log likelihood (reconstruction) terms for fitting the multi-view data; 
			\item[2).] The adversarial learning constraint that
			contribute to consistent latent encoding. 
		\end{enumerate*}
		The objective of our ACCA is given as } 
		\begin{eqnarray}\label{eq:ACCA_objective}
		\lefteqn{\min \limits_{\Theta,\Phi}\;\mathcal{L}_{\rm ACCA} ({x},{y}) = \frac{1}{N} \sum_{\{x,y\}} {\big[}-\mathbb{E}_{ {q_{\phi_{xy}}({z}|{x},{y})}}[\log {p_{\theta_x}({x}|{z})}+\log {p_{\theta_y}({y}|{z})}]}\\
		&&\qquad \qquad \qquad \qquad \qquad \quad \:-\mathbb{E}_{ q_{\phi_{x}}({z}|{x})}[\log {p_{\theta_x}({x}|{z})}+\log {p_{\theta_y}({y}|{z})}]\nonumber\\ 
		&&\qquad \qquad \qquad \qquad \qquad\qquad \:  - \mathbb{E}_{q_{\phi_{y}}({z}|{y})}[\log {p_{\theta_x}({x}|{z})}+\log {p_{\theta_y}({y}|{z})}] + \mathcal{R}_{\rm{GAN}}{\big]},\nonumber
		\end{eqnarray}
		where $\Theta$ and $\Phi$ denotes the parameters of the encoders and the decoders respectively, i.e. $\Theta$  = $\{\theta_{x},\theta_{y}\}$ and $\Phi$ = $\{\phi_{xy},\phi_{x},\phi_{y}\}$. 
		
		The ACCA framework, as illustrated in Fig.~\ref{fig:network}, consists of 6 subnetworks. The three encoders, $\{E_{x},E_{xy},E_{y}\}$, and the two decoders $\{D_{x},D_{y}\}$ constitute the view-reconstruction scheme, which correspond to the first three terms in Eq.~\eqref{eq:ACCA_objective}. The three encoders (generators), together with the shared discriminator $\hat{D}$, compose the adversarial distribution matching scheme. These subnetworks, i.e. $\{E_{x},E_{xy},E_{y},\hat{D}\}$ compose the adversarial regularizer that promote with $\mathcal{R}_{\rm GAN}$, namely
		\begin{align}
		\lefteqn{\mathcal{R}_{\rm GAN}(E_x,E_y,E_{xy},\hat{D})=\mathbb{E}_{{z}\sim p_({z})}\log(\hat{D}({z}))+\mathbb{E}_{{{z}_{{xy}}\sim q_{\phi_{xy}}({z}|{x,y})}}\log(1-\hat{D}({z}_{{xy}}))}
		\\
		&&\qquad\qquad\qquad\qquad\qquad+\mathbb{E}_{{z}_{{x}}\sim q_{\phi_{{x}}}({z}|{x})}\log (1-\hat{D}({z}_{{x}}))+ \mathbb{E}_{{z}_{{y}}\sim q_{\phi_{y}}({z}|{y})}\log (1-\hat{D}({z}_{{y}})). \nonumber
		\end{align}
		Here, we add the subscripts to discriminate the latent codes $z$ encoded from different views $X, Y, XY$. This distinctiveness is criterial in the experiment part.
		
		\begin{figure}[t]
			\centering
			\hspace{-5mm}\includegraphics[width=0.9\textwidth]{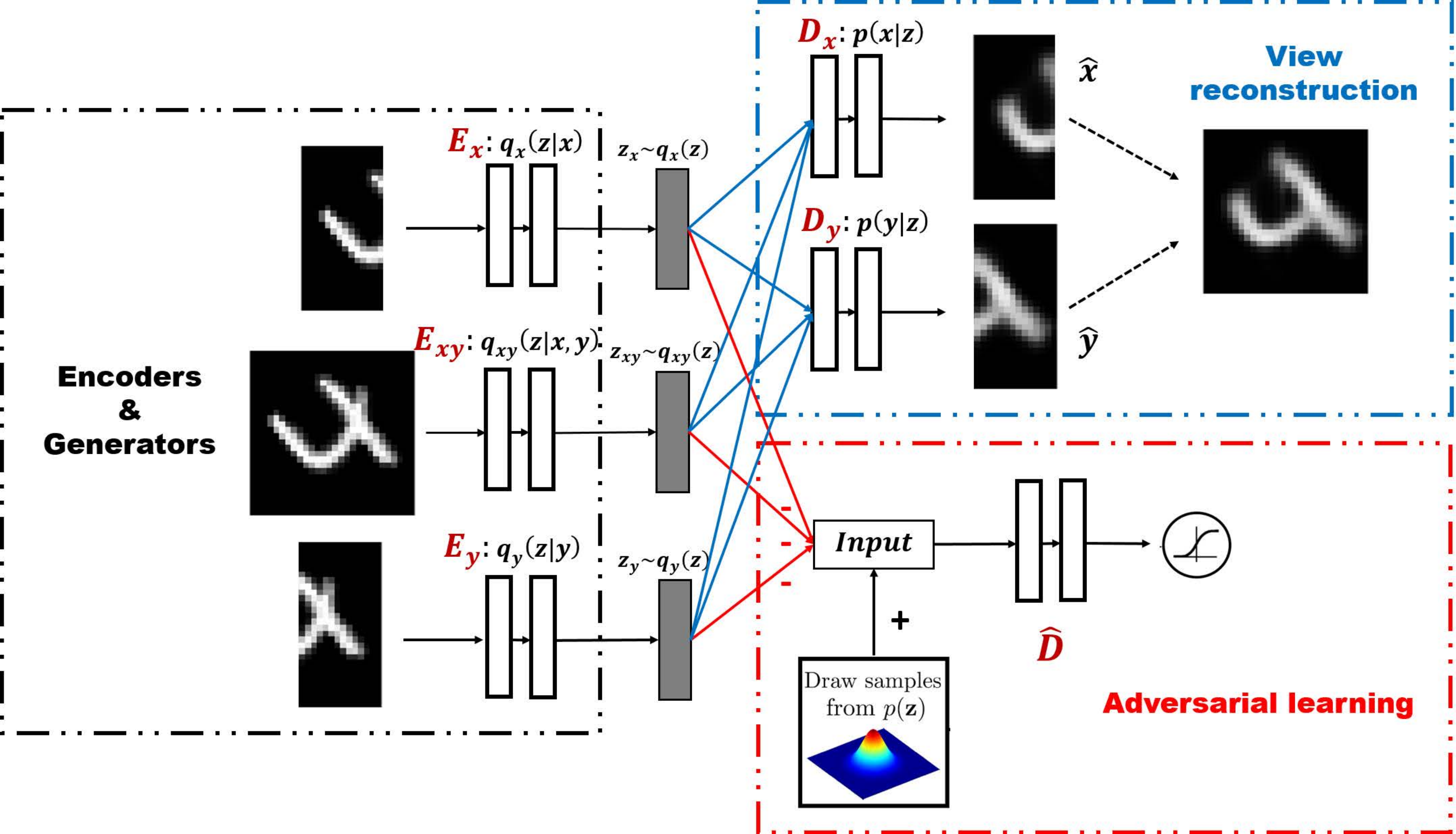}
			\caption{\label{fig:network}  Overall structure of ACCA. The left panel represents encoding with holistic information scheme; the top right panel corresponds to the cross-view reconstruction; the bottom right panel illustrates the adversarial learning criterion.}
		\end{figure}
		
		In practical, our ACCA is jointly trained by alternatively updating two phases- the reconstruction phase and the regularization phase. In the reconstruction phase, we update the encoders and the decoders to minimize the reconstruction error of the two principle views. In the regularization phase, the adversarial networks, with multiple encoders
		 or generators,
		 are trained following the same alternating procedure as in ~\cite{hoang2018mgan}. Once the training procedure is done, the encoders will define expressive encodings for each view.
		
		\section{Connection to other models}\label{sec:discussion}
		In this section, we discuss the connection between ACCA and other existing models.
		
		\subsection{{Understanding CCA models with CMI}}
	From a {Bayesian perspective}, the general CCA models come with an assumption that the two views, $X$ and $Y$, are conditionally independent given the latent variable $Z$, i.e. Eq.~\eqref{eq:conditional_independent}, to achieve a tractable solution for inference. However, such an assumption is hard to verify in real multi-view analysis problems that incorporate complex distributions. Here, we analyse this inherent assumption of CCA {with conditional mutual information}.
		
		Given random variables ${X}, {Y}$ and ${Z}$, the conditional mutual information~(CMI) defines the expected KL-divergence between the conditional joint distribution $p{(x,y|z)}$ and the product of the conditional marginal distributions, $p(x|z)$ and $p(y|z)$~\citep{zhang2014conditional}. 
		\begin{equation}\label{eq:CMI_KL_divergence}
		I{({X};{Y}|{Z})}= \mathbb{E}_{p(z)} [D_{KL}(p(x,y|{z})\parallel p({x}|{z})p({y}|{z}))] \geq 0 
		\end{equation} 
		The minimum, \begin{small}{ $I{({X};{Y}|{Z})}=0$}\end{small}, can only be achieved when $X$ and $Y$ are conditional independent given $Z$. Consequently, the conditional independent criterion of CCA (Eq.~\eqref{eq:conditional_independent}) can be achieved by minimizing the CMI. The objective can be given as
		\begin{eqnarray}\label{eq:CMI_derivation}
		&& I_{\theta}{({X};{Y}|{Z})} \nonumber \\ 
		&=& \iiint 
		p{({z})} p{({x},{y}|{z})} \log \frac{p{({x},{y}|{z})}}{p{({x}|{z})}p{({y}|{z})}} d{z}d{x}d{y} \nonumber \\
		&=& \iiint p{({z}|{x},{y})} p{({x},{y})} [\log \frac{p{({x},{y}|{z})}}{p{({x}|{z})}p{({y}|{z})}} -\log p({x},{y})+\log p({x},{y})] d{z}d{x}d{y} \nonumber \\
		&=& \begin{small}
		\iiint p{({z}|{x},{y})} p{({x},{y})} [\log \frac{p({z}|{x},{y})}{p({z})}-\log p({x}|{z})-\log p({y}|{z})+\log p({x},{y})]d{z}d{x}d{y}
		\end{small} \nonumber \\
		&=& H(X, Y)+\mathbb{E}_{p_{\theta}({x},{y})}[-\mathbb{E}_{p({z}|{x},{y})} [\log {p_{\theta}({x}|{z})}+\log {p_{\theta}({y}|{z})}] + D_{KL}(p_{\theta}({z}|{x},{y})\parallel p({z}))], \nonumber
		\end{eqnarray}
		where $H(X,Y)$ is a constant and has no effect on the optimization~\citep{gao2018auto}. Therefore, the minimum of CMI can be achieved by minimizing the remaining terms, namely 
		\begin{equation}\label{eq:CMI_objective}
		\min\limits_{\theta} \mathbb{E}_{p(x,y)}~[F_\theta(x,y)]  \simeq \frac{1}{N} \sum_{\{x,y\}} F_{\theta}(x,y),
		\end{equation}
		where $F_\theta(x,y)=-\mathbb{E}_{p_\theta({z}|{x},{y})}~[\log {p_\theta({x}|{z})}+\log {p_\theta({y}|{z})}] +D_{KL}(p_\theta({z}|{x},{y})\parallel p({z}))\label{eq:ICCA_CMI_Fp}.$
		
		Although Eq.~\eqref{eq:CMI_objective} presents an objective for minimizing CMI, it is hard to optimize since the posterior $p_\theta(z|x,y)$ is unknown or intractable for the practical multi-view learning problems. Consequently, existing methods make \textbf{\emph{different assumptions}} on the incorporated distributions, e.g. prior, likelihood, and posterior, and adopt approximate inference methods to achieve  tractable solutions for multi-view analysis.
		
		\vspace{1mm}
		\noindent\textbf{\emph{Example 1: PCCA}}~\citep{bach2005probabilistic}.~~With an explicit conditional independent assumption, PCCA adopts Gaussian assumptions for both the likelihood and the prior to achieve tractable solution for the inference in linear CCA. Under the conditional independent constraint, the minimum of CMI, i.e.~$I({X};{Y}|{Z})=0$, is naturally satisfied. Due to the conjugacy of the prior and the likelihood, the posterior in Eq.~\eqref{eq:ICCA_CMI_Fp} can be presented with an analytic solution, with which the model parameters can be directly estimated with EM algorithms. 
		\begin{equation}
		{z}\sim\mathcal{N}({0},{I}_{d}), \quad
		{x}|{z}\sim \mathcal{N}({W}_{x}{z}+ {\mu}_{x},{\Phi_{x}}),\quad
		{y}|{z}\sim\mathcal{N}({W}_{y}{z}+{\mu}_{y},{\Phi_{y}})\nonumber \label{eq:CMI_PCCA}
		\end{equation}
		\vspace{0.5mm} 
		\noindent\textbf{\emph{Example 2: MVAE}}~\citep{ngiam2011multimodal}. If we consider Gaussian models with \mbox{${z}\sim\mathcal{N}({\mu},{0})$}, $p_{\theta}(x|z) = \mathcal{N}(F_{\phi_{x}}(z_{x}), I)$ and $ p_{\theta}(y|z) = \mathcal{N}(G_{\phi_{y}}(z_{y}), I)$, the $z_{x}$ and $z_{y}$ are obtained as point embedding obtained with $F(x)$ and $G(y)$, i.e. $z_{x} = F_{\theta_{x}}(x)$ and $z_{y} = G_{\theta_{y}}(y)$~( Section~\ref{sec:vanillaCCA}). We can see that the reconstruction terms in Eq.~\eqref{eq:CMI_objective} measures the $l_{2}$ reconstruction error of the two inputs from the latent code $z$ through the DNNs defined with $F^{-1}$ and $G^{-1}$. {The objective of MVAE is}  
		\begin{eqnarray}
		\min \limits_{\theta,\phi} \frac{1}{2N} \sum_{\{x,y\}} ~ {\| x- F^{-1}_{\phi_{x}}(F_{\theta_{x}}(x))\|}^{2} + {\| y - G^{-1}_{\phi_{y}}(G_{\theta_{y}}(y))\|}^{2}. \nonumber \label{eq:CMI_MVAE}
		\vspace{-3mm}
		\end{eqnarray}
		{Note that, MVAE is a simple AE, with no regularization on posterior-and-prior matching.}
		
		\vspace{1mm}
		
		\noindent \textbf{\emph{Example 3: VCCA}}~\citep{wang2016deep}. Considering a model where the latent codes  ${z}\sim\mathcal{N}({\mu},\Sigma)$ and the observations $x|z$ and $y|z$ {both follow implicit distribution}, VCCA {adopts} variational inference to get the approximate posterior for~ Eq.~\eqref{eq:ICCA_CMI_Fp} with two additional assumptions: 1). The single input view can provide sufficient information for the multi-view encoding, {namely} $q_{\phi}(z|x,y)\approx {q_{\phi}(z|x)}$; 2). The variational approximate posterior $q_{\phi}(z|x) \sim \mathcal{N}(z;\mu,\Sigma)$, where $\Sigma = diag(\sigma_{1}^{2},\ldots ,\sigma_{d}^{2})$. In this case, the KL-divergence term can be explicitly computed with \mbox{$D_{KL}(q_{\phi}({z}|{x})\parallel p_{\theta}({z})) = -\frac{1}{2} \sum_{j = 1}^{d} (1-\sigma_{j}^{2}-\mu_{j}^{2}+ \log \sigma_{j}^{2})$}. Note that, $p_{0}(z)$ is defined with
		{explicit form} and the encoding functions actually models the distribution parameters. The latent codes is then obtained by sampling $L$ samples from the posterior distribution, i.e. ${z}^{l} \sim q_{\phi}({z}|{x})$, with the reparameterization trick~\citep{kingma2013auto}.  The objective of VCCA is given as 
		\begin{align}
		&\min \limits_{\theta,\phi} \frac{1}{N} \sum_{\{x,y\}}{\big[}-\frac{1}{L} \sum_{l = 1}^{L}[\log {p_{\theta}({x}|{z}^{l})}+\log {p_{\theta}({y}|{z}^{l})}] 
		+ {D_{KL}(q_{\phi}({z}|{x})\parallel p_{\theta}({z}))}{\big]}. \nonumber \\
		&\; s.t.\quad {z}^{l}_{x} = {\mu_{x}} + \Sigma_{x}\epsilon^{l},\ \text{where}\ \epsilon^{l} \sim\mathcal{N}({0},{I}_{d}),\ l= 1,\ldots,L. \label{eq:CMI_VCCA}
		\end{align}
		\textbf{\emph{Example 4: Bi-VCCA}}~\citep{wang2016deep}. Bi-VCCA adopts the encoding of both the two views, namely, $q_{\theta}({z}|{x})$ and $q_\theta({z}|{y})$ to approximate $q_\theta({z}|{x}, {y})$. Its objective is given as a {heuristic} combination of Eq.~\eqref{eq:CMI_VCCA} derived with each encodings, namely
		\begin{eqnarray}
		\lefteqn{\min \limits_{\theta,\phi} \frac{1}{N} \sum_{\{x,y\}}{\big[}[ -\frac{\lambda}{L} \sum_{l = 1}^{L}[\log {p_{\theta}({x}|{z}^{l}_{x})}+\log {p_{\theta}({y}|{z}^{l}_{x})}] 
			+{D_{KL}(q_{\phi}({z}|{x})\parallel p_{\theta}({z}))}]} \\
		&&\qquad \qquad \;\;+  [-\frac{1-\lambda}{L} \sum_{l = 1}^{L}[\log {p_{\theta}({x}|{z}^{l}_{y})}+\log {p_{\theta}({y}|{z}^{l}_{y})}] 
		+{D_{KL}(q_{\phi}({z}|{y})\parallel p_{\theta}({z}))}]{\big]}, \nonumber \\
		&&s.t.\quad {z}^{l}_{x} = {\mu_{x}} + \Sigma_{x}\epsilon^{l}, {z}^{l}_{y} = {\mu}_{y} + \Sigma_{y}\epsilon^{l},\ \text{where}\ \epsilon^{l} \sim\mathcal{N}({0},{I}_{d}),\ l= 1,\ldots,L,  \label{eq:CMI_BIVCCA}
		\end{eqnarray}
		where $\lambda \in [0,1]$ is the trade-off factor between the two encodings.
		
		\subsection{{ACCA versus existing CCA methods}}\label{sec: ACCA_superiority}
		
		Based on our analysis, we emphasise the superiority of the proposed ACCA (Fig.~\ref{fig:motivation}(c)) {over the aforementioned CCA prototypes} in the following aspects. 
		\begin{itemize}
			\item[1.] The adversarial learning criterion enables ACCA to achieve a tractable solution for multi-view analysis with much flexible prior and posterior distributions. This benefits the expressiveness of the obtained aligned latent space.
			\item[2.] The adversarial learning criterion leads to consistent latent encoding in ACCA by matching marginalization of the incorporated distributions and thus facilitates ACCA to achieve better instance-level alignment for the multiple views. 
			\item[3.] {Appending $q(z|x,y)$ with the auxiliary view $XY$, our ACCA can better estimate the minimizing CMI objective (Eq.~\eqref{eq:CMI_objective}), compared with other variants that simply adopt the encodings from individual views, i.e. $q(z|x)$ and $q(z|y)$.}
		\end{itemize}
	
	Some works adopt {\emph{additional penalties}}, e.g. sparsity constraint~\citep{shi2019label}, on the these prototypes to further enhance the multi-view alignment. For instance, ~\cite{kidron2007cross} extends classic CCA with sparsity to enhance its performance on cross-modal localization task. \cite{DBLP:conf/nips/JiaSD10} introduces structured sparsity into MVAE. \cite{virtanen2011bayesian} proposes a generative CCA variant that also adopts KL-divergence as the criterion, with only an additional group sparsity assumption to improve the variations approximation. Note that, we can also extend ACCA with corresponding structural priors to enhance the alignment of the multiple views~\citep{DBLP:conf/icml/MathieuRST19}.}
	
	Some other works extend these prototypes by further exploiting {\emph{view-specific information}}. Besides the multi-view shared information, these variants also considers specific information in each view to benefit the alignment task. As a representative, VCCA-private extends Bi-VCCA by introducing two hidden-variables, $h_{x}$ and $h_{y}$, to capture the private information that is not captured with the common variable, $Z$. It adopts two extra KL-divergence constraints to match the encoding of the private variables {(Eq.(10)) in~\citep{wang2016deep}}. Our ACCA can also be naturally extended with such private variables and additional discriminators to further enhance the alignment. 
	 
	There are also generative CCA works that incorporates {\emph{additional information}, e.g. supervision,} to benefit the multi-view alignment. For example, Multi-view Information Bottleneck (MVIB)~\citep{federici2020learning} aligns the two views in a supervised manner, in order to obtain multi-view data representations that are maximumly informative about the downstream prediction task, i.e. $X_1, X_2\rightarrow Y$. (Note that, $Y$ denotes the label here.) {Consequently, its motivation is different from our ACCA, which target at alignment of the two views for generation tasks, i.e. $X_1\leftrightarrow X_2$. Actually, MVIB  even cannot facilitate our targeted cross-view generation task due to the lack of a generation mechanism.} In addition, besides the minimum CMI criterion, Eq.~\eqref{eq:CMI_KL_divergence}, that formulates ACCA, MVIB also adopts an additional superfluous information minimization objective to discard the input information that is irrelevant to its label.
	{\small	\begin{equation}
		\mathcal{L}_{MIB}(\theta;\lambda) = \underbrace{I_{\theta}(Z; X_{1} | X_{2})}_{\text{superfluous information}} + \lambda I_{\theta}( X_{1};X_{2}| Z) \nonumber
		\end{equation}}
\vspace{-1mm}
\noindent{In this sense, MVIB can be regarded as an extension of multi-view alignment that further incorporates superfluous information to handle supervised downstream tasks. We can also apply our developed inference method in ACCA to solve the generative variant of the MVIB objective as well.}

{Note that, in this work, we focus on studying the classic CCA prototypes in terms of the multi-view alignment for data generation. Consequently, the above CCA variants with additional penalties, or with view-specific variables are not for main comparisons here. The MVIB is not comparable here since it even can not facilitate generation.}

		\subsection{{ACCA versus Adversarial Autoencoders~(AAEs)}}
		Another work that is highly relevant to our ACCA, is \textbf{\emph{Adversarial Autoencoders}}~(AAEs) ~\citep{makhzani2015adversarial}. AAEs adopt adversarial distribution matching to promote the reconstruction of autoencoders, {based on }Variational Autoencoders (VAEs). Compared with AAEs, our ACCA is contributive since we extend the adversarial distribution matching into the multi-view scenario to facilitate multi-view alignment, especially for cross-view generation tasks. We also elaborate that our model is reasonable to achieve superior alignment for multiple views with consistent latent encoding, by analysing the conditional independent assumption in CCA with CMI. 
		\subsection{{Instance-level alignment versus distribution-level alignment}}
		In this work, we study instance-level multi-view alignment with CCA, namely to achieve correspondence for instance-embeddings obtained from each view. There are also works that study the distribution-level alignment of the multiple views. These works focus on aligning the marginal distribution of the views, i.e. $P(X)$ and $P(Y)$ without considering the pairwise correspondence for each instance. Cross-view generation in such setting is regarded as a style transfer task~\citep{ganin2016domain}. For example, Cycle-GAN~\citep{zhu2017unpaired} studies unsupervised image translation in two domains by modelling cycle consistency. UNIT~\citep{DBLP:conf/nips/LiuBK17} and MUNIT~\citep{DBLP:conf/eccv/HuangLBK18} study the same task by incorporating a common latent space into Cycle-GAN. Conditional GANs are adopted to facilitate the cross-view image synthesis task~\citep{regmi2018cross}. As this is not the focus of our paper, we do not {discuss them further}.
		
		\section{Experiments}\label{sec:experiments}
	
		In this section, {we evaluate the performance of our ACCA regarding multi-view alignment and generation.} We first testify the advantages of ACCA in Section~\ref{sec:ACCA_verification}. Then, we show the superiority of ACCA in achieving multi-view alignment in three aspects. We conduct correlation analysis to show ACCA captures higher nonlinear correlation among the multiple views in Section~\ref{sec:correlation_analysis}. We present alignment verification to show ACCA achieves better instance-level correspondence in the latent space in Section~\ref{sec:alignment_verification}. We conduct several cross-view analysis tasks with noisy inputs to show the robustness of ACCA in achieving instance-level alignment of the multiple views in Section~\ref{sec:cross-view_generation}. 
		
		 {We also evaluate the quality of obtained embeddings regarding downstream supervised tasks, to demonstrate our ACCA facilitates superior alignment without sacrificing discriminative property of the representation. The experiments regarding clustering and classification are presented in Section~\ref{sec:exp_clustering} and Section~\ref{sec:exp_classification}, respectively.}
		 	
		{Note that, our work target at instance-level multi-view alignment and generation. Consequently, we emphasize the evaluation in the first few subsections, i.e. the preserved correspondence on the latent embeddings and how well the correspondence can be recovered from the obtained latent spaces, cross-view generation. {The evaluation of the discriminative property of latent embeddings is presented for better illustration.}}
		
		{In Section~\ref{sec:bivcca_private}, we present a preliminary study on the influence of view-specific variables for alignment and generation as future works.}
		
		\subsection{{Superiority of adversarial criterion for multi-view alignment}}\label{sec:ACCA_verification}
		
		We first testify the benefits achieved with the adversarial learning alignment criterion, i.e. consistently matching the marginalized latent encodings with flexible priors. 
		
		\subsubsection{{Consistent encoding in ACCA}}	
		We verify the consistent encoding in ACCA with one of the most commonly used multi-view learning dataset - MNIST left/right halved dataset (MNIST\_LR)~\citep{andrew2013deep}. Details about the dataset and network design are shown in Table~\ref{tab:dataset}.
		
		To testify the approximation of the three encodings in ACCA, we estimate the distribution distances among the three posterior distribution with kernel Maximum Mean Discrepancy (MMD)~\citep{DBLP:journals/jmlr/GrettonBRSS12}. Specially, we assign Gaussian mixture prior (Eq.~\eqref{eq:zprior}) for ACCA, and then calculate the sum of the MMD distance between the three encodings and the prior $p_{0}(z)$ in Eq.~\eqref{eq:marginalization} during the training process. Fig.~\ref{fig:consistent_encoding} shows that the distance gradually decreases during the convergence of ACCA. This trend verifies that ACCA can facilitate the matching of non-Gaussian marginalized posteriors, i.e. the consistent encoding (Eq.~\eqref{eq:ACCA_three_appro}).
		\begin{figure*}[t]
			\centering
			\hspace{-1mm}
			\subfigure{\includegraphics[width=0.50\textwidth]{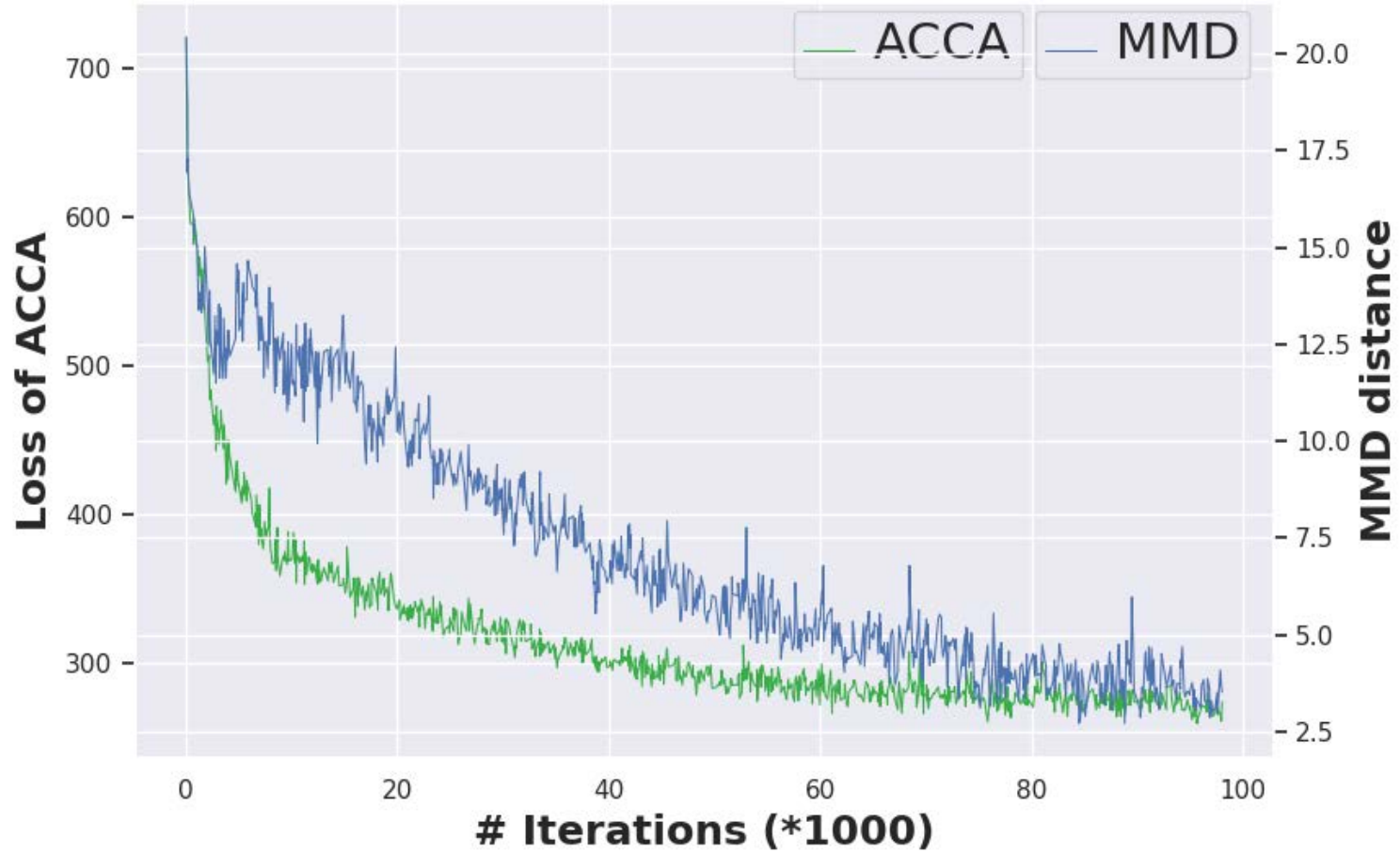}}
			\subfigure{\includegraphics[width=0.492\textwidth]{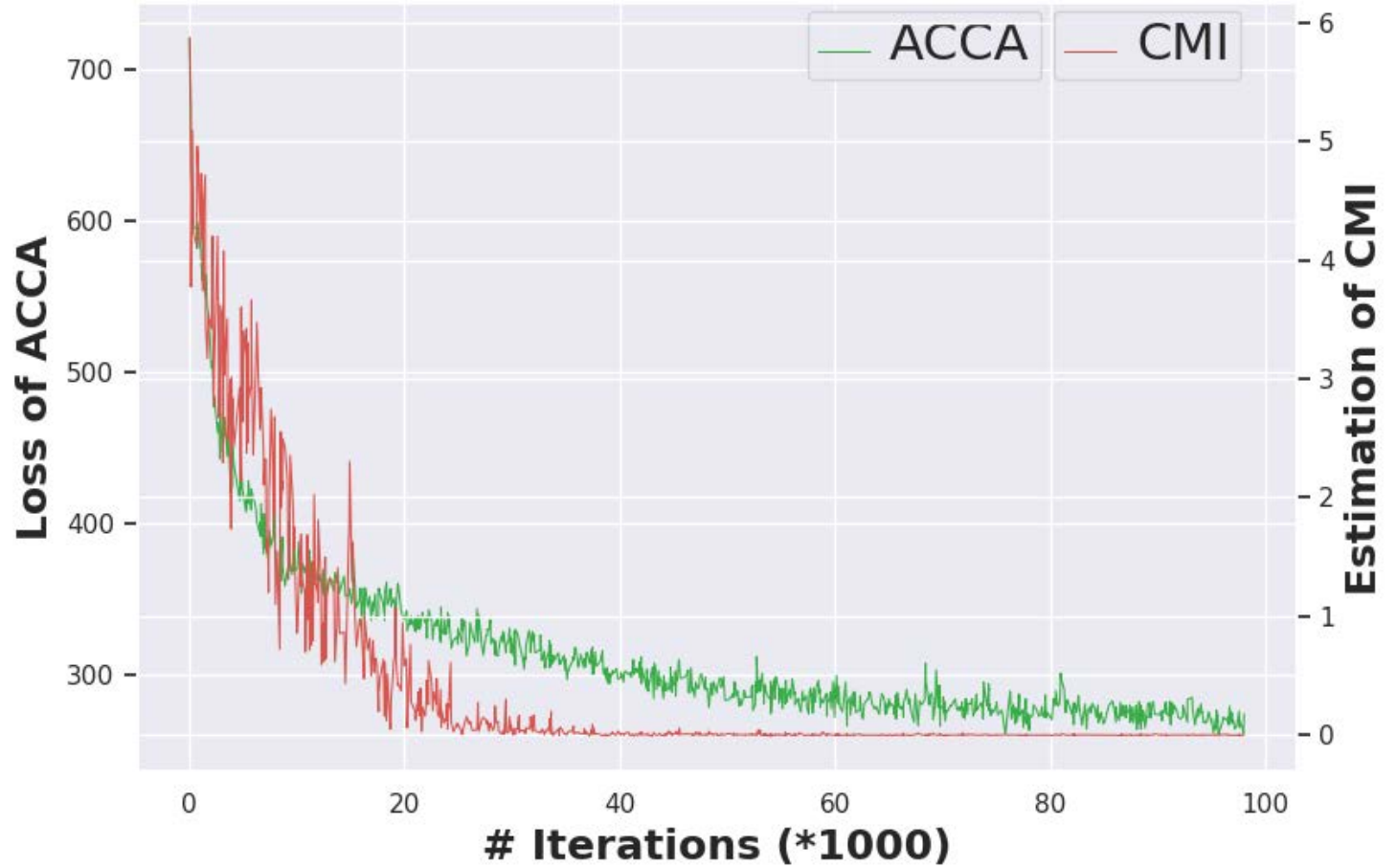}}
			\caption{Verification of consistent encoding in ACCA. \textbf{Left:} The holistic encodings are approximated, i.e. Eq.~\eqref{eq:ACCA_three_appro}, during the training of ACCA. \textbf{Right:} The minimum CMI, i.e. Eq.~\eqref{eq:CMI_KL_divergence}, is implicitly achieved in ACCA.}\label{fig:consistent_encoding}
			\vspace{-2mm}
		\end{figure*}
		
		We also estimate CMI during the model training process with an open-source non-parametric Entropy Estimation toolbox\footnote{https://github.com/gregversteeg/NPEET}. The right subfigure of Fig.~\ref{fig:consistent_encoding} illustrates that the CMI gradually decreases during the training of ACCA and it reaches to zero at a relatively early stage in the convergence of ACCA. The trend indicates that ACCA implicitly minimizes CMI and the optimal, $I{({X};{Y}|{Z})=0}$, can be achieved at its convergence. Consequently, the explicit conditional independent constraint (Eq.~\eqref{eq:conditional_independent}) of CCA can be automatically satisfied in our ACCA.
		
		\subsubsection{{Flexibility of prior encoding in alignment}}
		
		We conduct correlation analysis on a toy dataset with non-Gaussian prior to verify that ACCA benefits from handling implicit distributions for multi-view alignment.
		
		\vspace{1mm}
		
		\noindent \textbf{Toy dataset:} Following ~\cite{DBLP:journals/nn/Hsieh00}, we construct a toy dataset that exists nonlinear dependency between the two views for testing. Let $X = W_{1}Z$ and $Y = W_{2}Z^{T}Z$, where $Z$ denotes a 10-D vector with each dimension $z\sim p(z)$, and $W_{1}\in \mathbb{R}^{10\times50}$ , $W_{2}\in \mathbb{R}^{10\times50}$ are the random projection matrices to construct the data. Details for the setting are presented in Table~\ref{tab:dataset}. As we consider nonlinear dependency with non-Gaussian prior, we set $p_{0}({z})$ with a mixture of Gaussian distribution in this experiment.
		\begin{equation}
		z \sim p(z) = 0.2 \times \mathcal{N}(0,\,1) + 0.5 \times \mathcal{N}(8,\,2) + 0.3 \times \mathcal{N}(3,\,1.5).\label{eq:zprior}
		\end{equation}
		
		\noindent\textbf{Dependency metric:} Hilbert Schmidt Independence Criterion (HSIC)~\citep{DBLP:conf/alt/GrettonBSS05} is a commonly used measurement for the overall dependency among variables. In this work, we adopt the normalized estimate of HSIC(nHSIC)~\citep{wu2018dependency} as the metric to measure the dependency captured by the embeddings of the test set ($Z_{X_{Te}}$ and $Z_{Y_{Te}}$) of each method. We report the nHSIC computed with both the linear kernel and the RBF kernel ($\sigma$ is set with the F-H distance between the points).
		\newcommand{\tabincell}[2]{\begin{tabular}{@{}#1@{}}#2\end{tabular}} 
		\begin{table}[t]
			\centering
			\large  
			\caption{ Details of the datasets and network settings with MLPs.
			} \label{tab:dataset}
			\vspace{1mm}
			\setlength{\tabcolsep}{1.2mm}{
				\scalebox{0.6}{%
					\renewcommand{\arraystretch}{1.2}
					\begin{tabular}{|c|c|c|c|c|}
						\hline
						Dataset & Statistics & \tabincell{c}{\tabincell{c}{Dimension\\ of ${z}$}} & \tabincell{c}{Network setting (MLPs) \\ $\hat{D}= \{1024,1024,1024\}$} & Parameters \\ \hline
						\tabincell{c}{Toy dataset \\ (Simulated)} & \tabincell{c}{\# Tr= 8,000 \\ \# Te= 2,000} & d = 10 & \tabincell{c}{$E_{x}= \{1024,1024\}$; \\ $E_{xy}=\{1024,1024\}$; \\ $E_{y}=  \{1024,1024\}$}
						& \multirow{5}{*}{\tabincell{c}{ \\ \\ \\For all the dataset: \\ learning rate = 0.001, \\  epoch = 100. \\ \\  For each dataset: \\ batch size tuned over \\ $\{16, 32, 128, 256, 500, 512, 1000\}$;\\ $d$ tuned over $\{10, 30, 50, 100\}$}} \\ \cline{1-4}
						\tabincell{c}{MNIST L/R halved dataset \\ (MNIST\_LR) \\ \citep{andrew2013deep}} & \tabincell{c}{\# Tr= 60,000 \\ \# Te= 10,000} &  d = 30 & \tabincell{c}{$E_{x}= \{2308,1024,1024\}$; \\ $E_{xy}=\{3916,1024,1024\}$; \\ $E_{y}=  \{1608,1024,1024\}$} & \\ \cline{1-4}
						\tabincell{c}{MNIST noisy dataset \\ (MNIST\_Noisy) \\ \citep{wang2016deep}} & \tabincell{c}{\# Tr= 60,000 \\ \# Te= 10,000} & d = 50 & \tabincell{c}{$E_{x}= \{1024,1024,1024\}$; \\ $E_{xy}=\{1024,1024,1024\}$; \\ $E_{x}=  \{1024,1024,1024\}$} & \\ \cline{1-4}
						\tabincell{c}{Wisconsin X-ray \\ Microbeam Database \\ (XRMB) \\ \citep{wang2016deep}} & \tabincell{c}{\# Tr= 1.4M \\ \# Te= 0.1M} & d = 112 & \tabincell{c}{$E_x= \{1811,1811\}$; \\ $E_{xy}=\{3091,3091\}$; \\ $E_y=  \{1280,1280\}$} & \\ \cline{1-4}  \hline
			\end{tabular}}}
			\vspace{-2mm}
		\end{table}
	
		\vspace{1.5mm}
		\noindent \textbf{Baselines:} We compare ACCA with several state-of-the-art vanilla CCA variants here.
		\begin{itemize}
			\item { \textbf{CCA}~\citep{hotelling1936relations}: Linear CCA model that learns linear projections of the two views that are maximally correlated.}
			\item { \textbf{PCCA}~\citep{bach2005probabilistic}: Probabilistic variant of linear CCA. }
			\item { \textbf{DCCA}~\citep{andrew2013deep}: DeepCCA, nonlinear CCA extension with DNN. }
			\item { \textbf{MVAE}~\citep{ngiam2011multimodal}: Multi-View AutoEncoders, an CCA variant that discovers the dependency among the data via multi-view reconstruction.}
			\item { \textbf{Bi-VCCA}~\citep{wang2016deep}: Bi-deep Variational CCA, a representative generative nonlinear CCA model restricted with Gaussian prior.}
			\item { \textbf{ACCA\_NoCV}: {An variant of ACCA which is designed without the encoding of the complementary view $XY$. This is used to verify the efficiency of the holistic encoding scheme in ACCA.}}
			\item { \textbf{ACCA(G)}; ACCA implemented with the standard Gaussian prior.}
			\item { \textbf{ACCA(GM)}: ACCA implemented with the exact Gaussian mixture prior.}
		\end{itemize}
		Since ACCA handles posterior distributions implicitly, its latent space can be more expressive to reveal the correspondences of the multiple views, compared with other baselines that can only directly handle simple Gaussian priors. {(An additional sampling procedure is requested for these methods to handle other complex distributions.)} Consequently, higher nonlinear dependency is expected to achieve in ACCA, especially when given the exact prior of the multi-view dataset. Table~\ref{hsic} reports the dependency captured in the common latent space of each method. The results are revealing in several ways: 
		\begin{itemize}
			\item[1).] Both CCA and PCCA achieve low nHSIC value on the toy dataset, due to their insufficiency in capturing nonlinear dependency.
			\item[2).] {DCCA achieves higher HSIC scores compared with other baselines due to its objective, which directly targets at higher linear correlations. However, its result is still inferior to all of our methods. }
			\item[3).] The results of MVAE and Bi-VCCA are unsatisfactory. The results of MVAE are not good, because it lacks the inference mechanism to qualify the encodings. Bi-VCCA gets inferior results mainly because of the inconsistent encoding problem caused by the inferior alignment criterion. 
			\item[4).] Our ACCA model all achieve good performance here. This indicates that the consistent encoding imposed by the adversarial distribution matching benefits the models' ability to capture nonlinear dependency.
			\item[5).] ACCA (GM) archives the best result in both settings. This verifies that ACCA benefits from the ability to handle implicit distributions.
		\end{itemize}
	
		\begin{table}[t]
		\centering
		\renewcommand{\arraystretch}{1.4}
		\caption{The dependency (higher is better) of latent embeddings. The best are in bold.} 
		\hspace{-3mm}\vspace{-1mm}
		\label{hsic}
		\scalebox{0.66}{
			\begin{tabular}{c|c|ccccc|ccc}
				\toprule
				\textbf{Metric}&\textbf{Datasets} & \textbf{CCA} & \textbf{PCCA} & \textbf{DCCA}& \textbf{MVAE} & \textbf{Bi-VCCA} &\textbf{ACCA\_NoCV} & \textbf{ACCA (G)} & \textbf{ACCA (GM)} \\ \hline
				\multirow{5}{*}{{\begin{tabular}[c]{@{}c@{}} \textbf{nHSIC}\\(linear kernel)\end{tabular}}}
				&toy        &0.0010   &0.1037&0.5353 &0.1428& 0.1035&0. 8563&  0.7296& 
				\textbf{0.9595}\\ 
				&MNIST\_LR  & 0.4210  &0.3777&0.6699&0.2500&0.4612 & 0.5233 & 0.5423 & \textbf{0.6823}\\
				&MNIST\_Noisy&0.0817 &0.1037&0.1460& 0.4089 &0.1912 &0.3343 & 0.3285& \textbf{0.4133}\\
				&XRMB      &  0.0574 &0.0416&0.2970& 0.2637&  0.1046& 0.1244 & 0.2903& \textbf{0.3482}\\
				&Maps       &-&-&0.3465& 0.4423& 0.1993 & 0.7324 &0.5157& \textbf{0.7043}\\ \hline
				\multirow{5}{*}{{\begin{tabular}[c]{@{}c@{}} \textbf{nHSIC}\\(RBF kernel)\end{tabular}}}
				&toy         &0.0029 &0.2037&0.7685& 0.2358 & 0.2543 & 0.8737 & 0.5870 & 
				\textbf{0.8764}\\ 
				&MNIST\_LR   & 0.4416   &0.3568&0.6877& 0.1499  &0.3804& 0.5799 & 0.6318 & \textbf{0.7387}\\
				&MNIST\_Noisy & 0.0948  &0.0993&0.1605& 0.4133&0.2076 &0.2697 & 0.3099& \textbf{0.4326}\\
				&XRMB       &  0.0534  &0.03184& \textbf{0.3180} & 0.0224 &  0.0846  & 0.1456 &0.2502 & 0.2989\\
				&Maps       &-&-&0.5905& 0.5624 & 0.3956 &0.8171 & 0.6285& \textbf{0.8658}\\
				\bottomrule
		\end{tabular}}
		\vspace{-1mm}
	\end{table}
		\subsection{{Correlation Analysis}}\label{sec:correlation_analysis}
		We further conduct correlation analysis on four commonly used multi-view datasets to testify the alignment achieved with each method. Higher correlations are expected with latent embeddings that preserve better data correspondence. Details about the datasets are presented in Table~\ref{tab:dataset} and Table~\ref{tab:dataset_generation}. {For XRMB, we follow the setting of DCCA, \citep{wang2016deep} --- we divide the data set into 5-folds, and report the average nHSIC scores for comparison. For ACCA (GM), we adopt the same prior as the Toy dataset, i.e.~\ref{eq:zprior}, as a simple arbitrary selection of non-Gaussian prior.} The results are presented in Table~\ref{hsic}. We can see that 
		
		\begin{itemize} 
			\item[1).]  {DCCA achieves higher correlation compared with the baselines that do not support data generation, i.e. CCA and PCCA. This is \mbox{because it adopts nonlinear} mapping, thus enable it to exploit nonlinear correlations in the input for alignment.}
			
			\item[2).] {The correlation achieved with MVAE is inferior to DCCA in most of the settings. This is because, MVAE seeks for embeddings that result in better view reconstruction. However, DCCA directly targets at the embeddings that achieve maximum linear correlation, which is generally coherent with the evaluation. }
			
			\item [3).] {Our methods, ACCA\_NoCV and ACCA, outperforms Bi-VCCA in all the settings. Our results are comparable and even better than DCCA in some of the settings. This indicates that our consistent encoding design can benefit the consistency preserved in the latent space. Since ACCA can facilitate data generation compared with DCCA, the comparison between ACCA and DCCA, MVAE indicates that ACCA can balance the data correspondence and reconstruction quality. The argument is collaboratively supported by the data generation result in Section~\ref{sec:cross-view_generation}.}

			\item[4).] {Among our three ACCA variants, the ACCA (GM) archives the best result almost all of the settings. This observation indicates that the preserved latent correspondence can be enhanced by incorporating a more expressive latent space with more flexible priors. It also verifies the superiority of our ACCA for directly handle flexible prior without extra sampling procedure. (Section~\ref{sec: ACCA_superiority}).} 
		\end{itemize} 

{In addition to the quantitative correlation analysis, we further conduct t-SNE visualization to demonstrate the quality of obtained embeddings. Specifically, in Fig.~\ref{fig:tsne_mnist}, we compare the embeddings of the two individual views obtained with DCCA, MVAE, BI-VCCA, and ACCA(G). It is clear that for the two vanilla CCA models, DCCA and MVAE, embeddings of each view fail to preserve distinguishable clustering structure. This observation can be explained with our analysis that they lack the inference mechanism to qualify the obtained embeddings. For Bi-VCCA, the embedding of view $X$ presents great clustering structure. But the embeddings of view $Y$ is disorderly scattered in the common latent space. This implies that the instances do not prohibit desired correspondence in the latent space, meaning that the two views are not well aligned with Bi-VCCA. The observation also implies that the left part of MNIST data potentially preserves more label information than the right views. For our ACCA, embeddings of both of the two views present good clustering structure. This indicates that the two views are better aligned with the proposed ACCA.} 

\begin{figure}[t]
	\centering
	\hspace{-4mm}
	\includegraphics[width=0.485\textwidth]{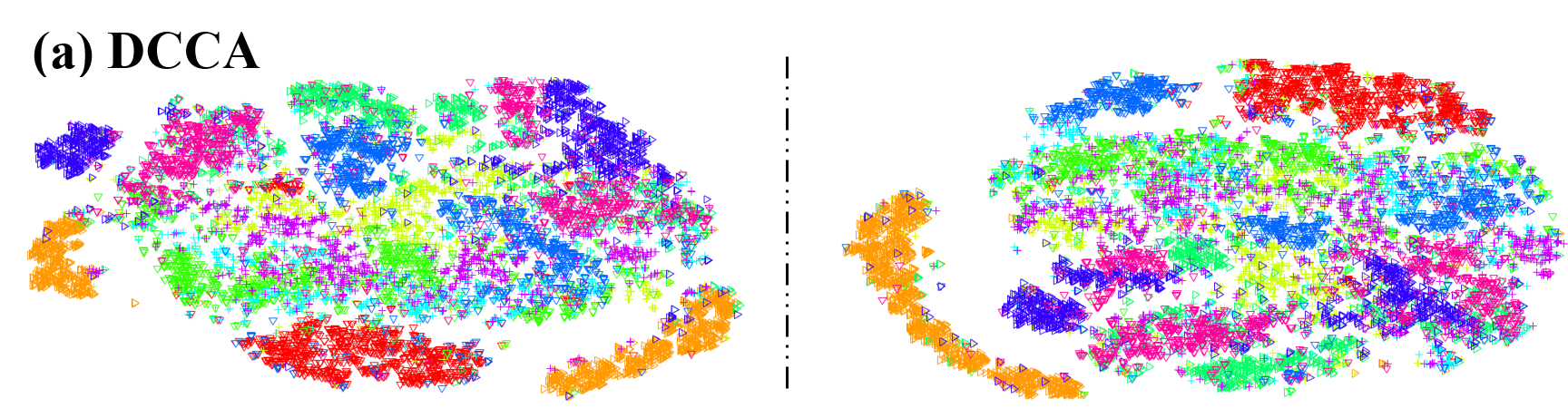}\hspace{3mm} \includegraphics[width=0.485\textwidth]{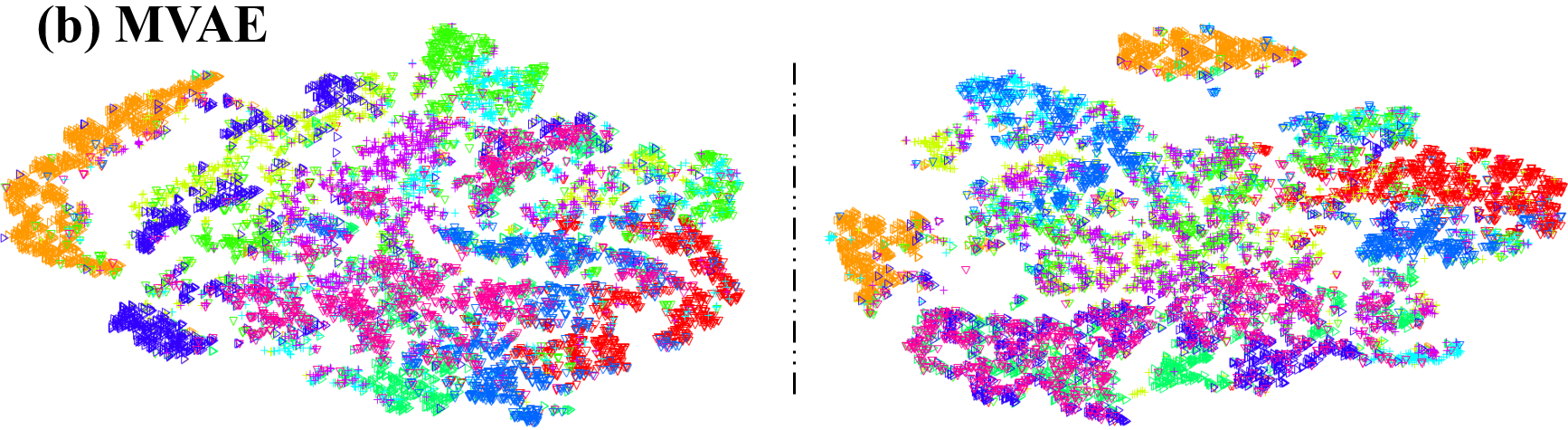} 
	\vspace{3mm}
	\includegraphics[width=0.485\textwidth]{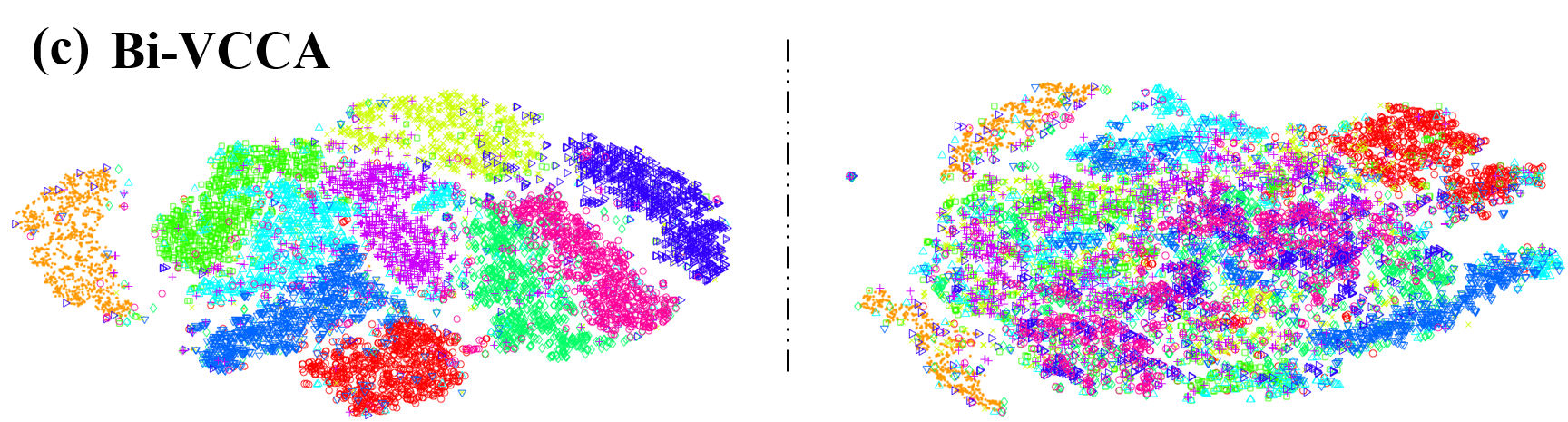}\hspace{3mm} 
	\includegraphics[width=0.485\textwidth]{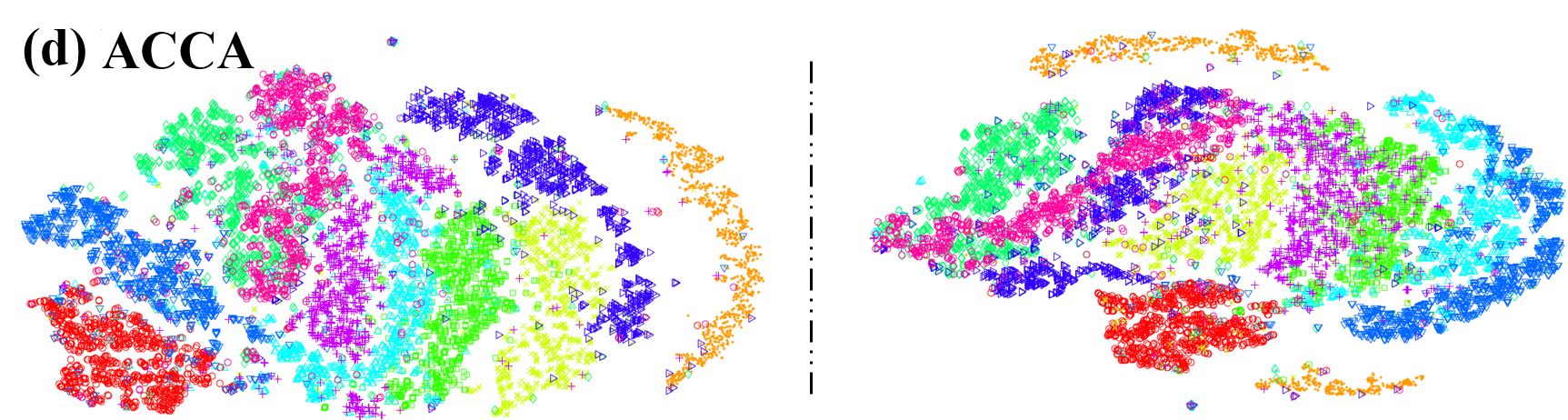} 
	\caption{\label{fig:tsne_mnist}  {t-SNE visualization of the embeddings of $X$[left] and $Y$[right] for MNIST\_LR, obtained with DCCA, MVAE, Bi-VCCA and ACCA, respectively. \mbox{The color represents label information.}}}
	\vspace{-3mm}
\end{figure}

\subsection{{Alignment verification}}\label{sec:alignment_verification}

We conduct alignment verification evaluate the instance-level correspondence achieved in the common latent space of ACCA. 
Specifically, we project the paired testing data of the \textbf{\emph{MNIST\_LR dataset}} to a two-dimensional latent space with Gaussian prior. We define \textbf{\emph{misalignment degree}} as the metric for the alignment performance. We take the origin point $O$ as the reference and adopt angular difference to measure the distance of the paired embeddings, i.e. $\phi ({z}_{x},{z}_{y}) = \angle {z}_{x}O{z}_{y}$. The misalignment degree of the multi-view is given by
		\begin{equation} \label{eq:alignment_metric}
		\delta = \frac{1}{N}\sum_{\{x,y\}}\frac{\psi({z}_{x},{z}_{y})} {\Psi},   
		\end{equation}
		where $N$ denotes the number of data pairs and $\Psi$ is the maximum angle among the paired embeddings (Fig.~\ref{fig:alignment}(d)). We compare ACCA with {DCCA,} MVAE, Bi-VCCA and ACCA\_NoCV here, since they are \textbf{\emph{baselines}} that have encodings for both the two views.
		
		The results are presented in Fig.~\ref{fig:alignment}. We have the following observations.
		\begin{itemize} 
			\item[1).] {For DCCA, the latent embeddings of two views are clearly scattered apart, indicating inferior instance correspondence in the latent space.  }
			\item[2).] The regions for the paired embeddings of Bi-VCCA are even not overlapped, and the misalignment degree of Bi-VCCA is $\delta = 2.3182$, which is much higher than the others. This indicates that Bi-VCCA severely suffers from the misaligned encoding problem.
			\item[3).] ACCA and ACCA\_NoCV, achieve superior alignment performance compared with DCCA, MVAE and Bi-VCCA. This shows the effectiveness of the consistent constraint on the marginalization for view alignment in ACCA. 
			\item[4).] The embeddings of ACCA are uniformly distributed in the latent space compared with ACCA\_NoCV. This indicates that the complementary view, $XY$ provide additional information for the holistic encoding. 
		\end{itemize} 
\begin{figure}[t]
	\centering
	\hspace{-4mm}
	\includegraphics[width=\textwidth]{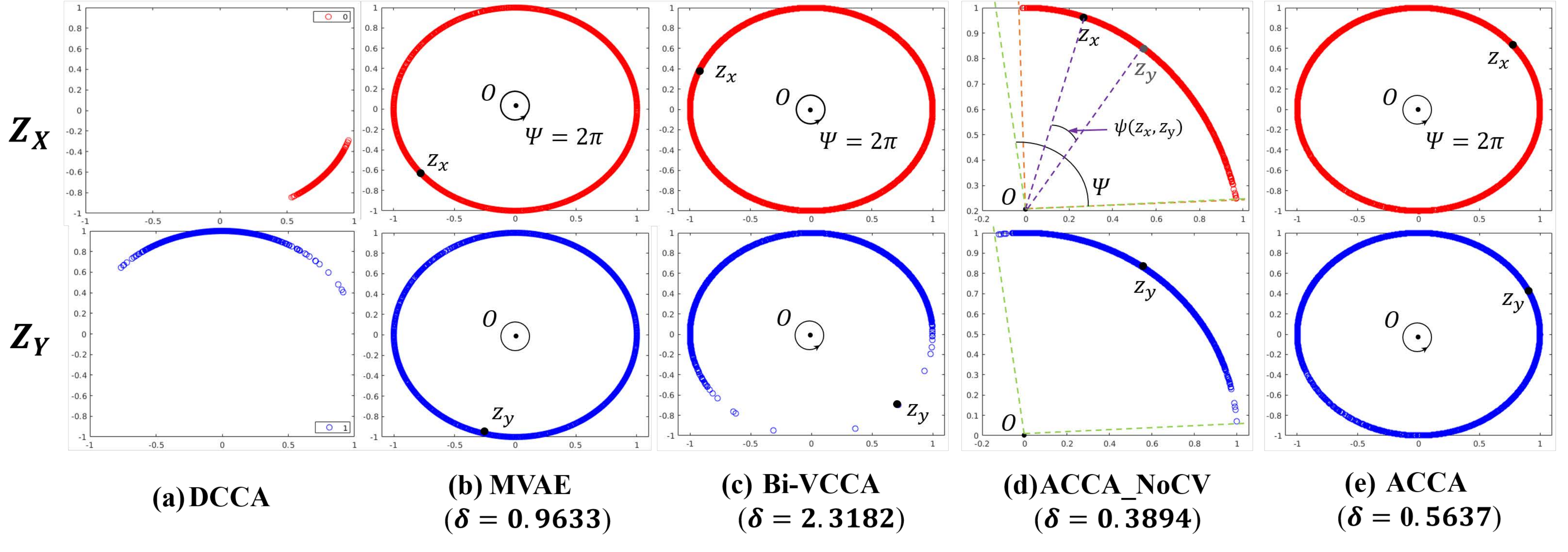} \vspace{-3mm}
	\caption{\label{fig:alignment}  Visualization of the embeddings obtained for the two views. Each row represents the embeddings obtained with view $X$ and view $Y$, respectively. (${z}_{x}$, ${z}_{y}$) denotes a pair of correspondent embedding. $\delta$ indicates the misalignment degree of each method. Methods with a smaller value of $\delta$ are better.}
	\vspace{-2mm}
\end{figure}

		\subsection{{Applications of cross-view generation}}\label{sec:cross-view_generation}
		
		We design several cross-view generation tasks to reflect the superior multi-view alignment achieved in ACCA. We first apply ACCA to image recovery task to conduct whole-image recovery, given the partial images as input for one of the views. We then test ACCA with face alignment task to annotate facial landmarks given the face images. Since MVAE and Bi-VCCA are the \textbf{\emph{baseline}} models that can support cross-view generation, we compare these two methods. We do not compare ACCA\_NoCV here since it is a variant of our ACCA and will be comparable with the ACCA due to the consistent encoding. We adopt Gaussian prior for ACCA here to conduct a fair comparison.
		
		\subsubsection{Image recovery}\label{sec:image_recovery}
		
		We testify the image recovery~\citep{sohn2015learning} performance of ACCA on \textbf{\emph{MNIST}} handwritten digit dataset and CelebFaces Attributes dataset (\textbf{\emph{CelebA}})~\citep{liu2015deep}. These two are both commonly used image generation datasets. The performance is evaluated based on the quality of generated images, e.g. is the image blurred? Does the image show apparent misalignment at the junctions in the middle? 
		
		\noindent\textbf{Image recovery on handwritten digits:} 
		We train the models with original data, while adding noise to the test data of MNIST dataset, to testify the robustness of the alignment achieved with each model. We divide the test data in each view into four quadrants and masked one, two or three quadrants of the input with grey color~\citep{sohn2015learning} and use the noisy images as the input for testing. The experimental result is evaluated from both qualitative and quantitative aspects.
		
		\vspace{1mm}
		\begin{figure}[t]
			\centering
			\includegraphics[width=0.85\textwidth]{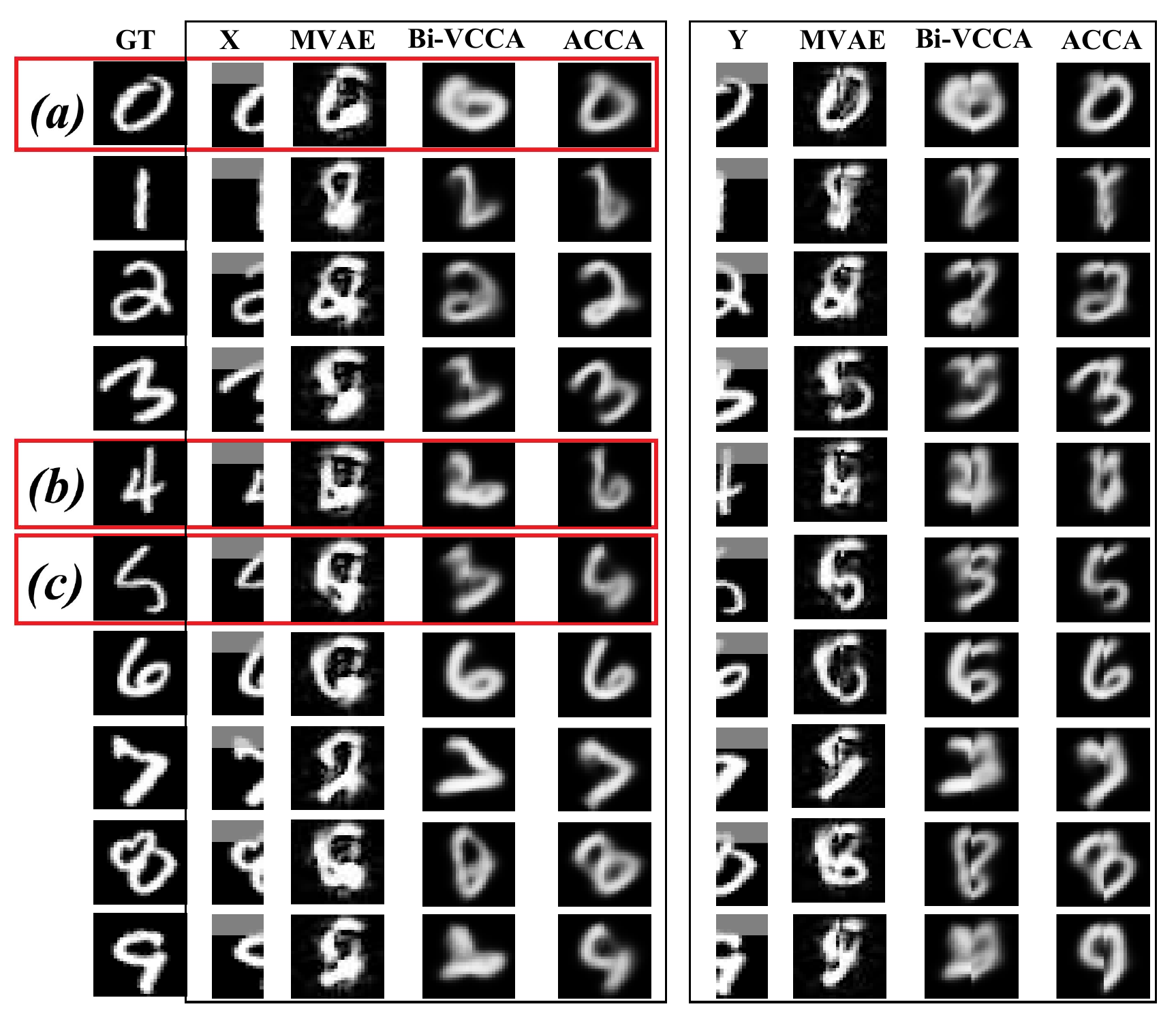}
			\vspace{-4mm}
			\caption{Generated samples given one quadrant noisy image as input. The first column is the ground truth. The next three columns show the input for view $X$ and the generated image with Bi-VCCA and ACCA, respectively. The last three columns are that of $Y$.}\label{fig:mnist_generation_result}
			\vspace{-3mm}
		\end{figure}
		
		\textbf{\emph{Qualitative analysis}}:
		Fig.~\ref{fig:mnist_generation_result} presents some of the recovered images (column 3-5) obtained with one-quadrant input. This figure clearly illustrates that, given the noisy input, the images generated with ACCA is more real and recognizable than that of MVAE and Bi-VCCA.  
	
\begin{table}
	\centering
	\renewcommand{\arraystretch}{1.25}
	\vspace{2mm}
	\caption{\small Pixel-level accuracy for image recovery with noisy inputs on the MNIST dataset.}
	\vspace{3mm}
	\label{tab:pixel-level-accuracy}
	\scalebox{0.9}{
		\begin{tabular}{c|c|ccc}
			\toprule
			\multirow{2}{*}{\begin{tabular}[c]{@{}c@{}}\textbf{Input}\\ (halved image)\end{tabular}} & \multirow{2}{*}{\textbf{Methods}}& \multicolumn{3}{c}{\textbf{Gray color overlaid}} \\ \cline{3-5} 
			& & 1 quadrant & 2 quadrants & 3 quadrants \\ \hline
			\multirow{3}{*}{\textbf{Left}} & MVAE & 64.94 & 61.81 & 56.15\\ 
			& Bi-VCCA & 73.14 & 69.29 & 63.05 \\ 
			& ACCA & \textbf{77.66} & \textbf{72.91} & \textbf{67.08} \\ \hline
			\multirow{3}{*}{\textbf{Right}} & MVAE & 73.57 & 67.57 & 59.69 \\ 
			& Bi-VCCA & 75.66 & 69.72 & 65.52 \\
			& ACCA & \textbf{80.16} & \textbf{74.60} & \textbf{66.80} \\ 
			\bottomrule
	\end{tabular}}
\end{table}
		\begin{itemize}
			\vspace{-1mm}
			\item [1).] The image generated with MVAE shows the worst quality. The images contain much more noise compared with other methods. In many cases, the ``digit'' is hard to identify, e.g. case (b). In addition, the generated image of MVAE shows clear misalignment at the junctions of the halved images, e.g. case (a).
			\item [2).] The images generated by Bi-VCCA are much more blurred and less recognizable than that of ACCA, especially in case (a) and case (b).
			\item[ 3).] ACCA can successfully recover the noisy half images, which are even confusing for our human to recognize. For example, in case (b), the left-half image of digit ``5'' looks similar to the digit ``4'', ACCA succeeds in recovering the true digit.
		\end{itemize}
		\textbf{\emph{Quantitative evidence}}: We compare the \textbf{\emph{pixel-level accuracy}} with the root mean \mbox{square error (RMSE), i.e.$1-RMSE$}. The results in Table~\ref{tab:pixel-level-accuracy} show that our ACCA consistently outperforms Bi-VCCA given the different level of masked input images. It is interesting to note that the whole images generated with the left-half images tends to be more realistic than that generated using the right-half. An probable reason is that the right-half images contain more information than the left-half images. {This finding coincides with our discovery in Fig.~\ref{fig:tsne_mnist}.{b}.} {This imbalance of information between the two views would drive the decoder of the less informative view to generate high-quality images, while sacrifice the alignment with another view.}
		\begin{figure}[t]
			\centering
			\hspace{-4mm}
			\includegraphics[width=\textwidth]{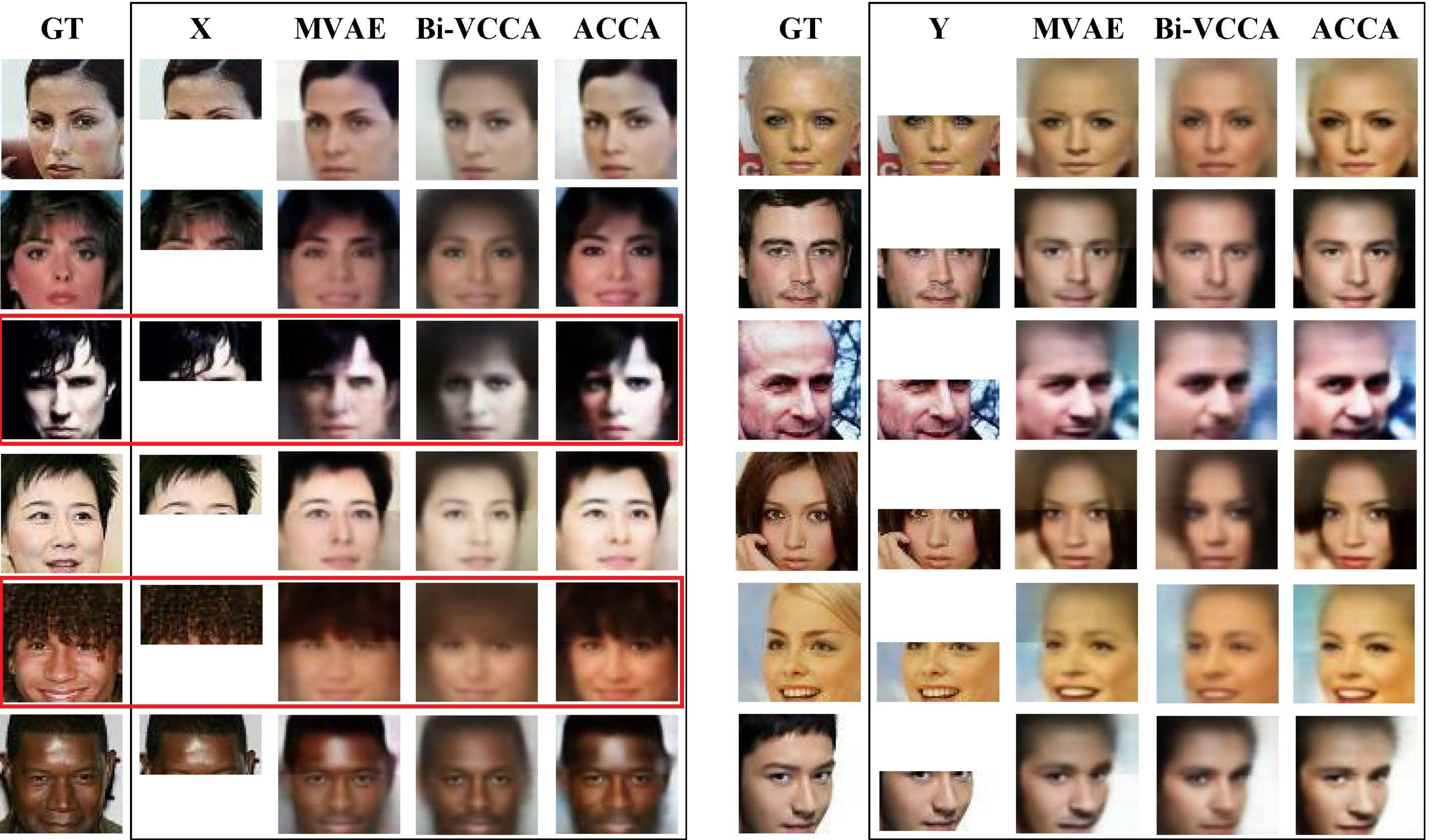}
			\vspace{-2mm}
			\caption{The image generated with different methods on CelebA. }\label{fig:face_generation_result}
			\vspace{-2mm}
		\end{figure}
	
	\vspace{1mm}

		\noindent\textbf{Image recovery on human faces:} For the human face recovery on the CelebA dataset, we halve the RGB images into top-half and bottom-half and design a CNN architecture to handle this task. Details of the network design are reported in Table~\ref{tab:dataset_generation}. 
		
		\vspace{1mm}
		
		\textbf{\emph{Qualitative analysis}}: Fig.~\ref{fig:face_generation_result} shows the image samples recovered for the CelebA dataset. We have mainly two observations.
		\begin{itemize}
			\item [1).] The samples generated by MVAE show clear misalignment at the junctions, especially when the images are with colored backgrounds. Some of the images are too blurred to see the details, e.g. the samples circled with red.
			\item [2).] The samples generated by Bi-VCCA are generally blurred than the other two. The observation is quite obvious in the image generated with the top-half image, which contains much fewer details than the bottom-half image.
			\item [3).] The images generated by ACCA show better quality compared with the others, considering both the clarity and the alignment of junctions.
		\end{itemize}
		
		\textbf{\emph{Quantitative evidence}}: We quantitatively assets the quality of generated images with the Frechet Inception Distance (\textbf{\emph{FID}})~\citep{heusel2017gans} and estimate the \textbf{\emph{sharpness}} of the generated test images using the image gradients~\footnote{We evaluate the sharpness of each test images using the gradients and average these values over the 1000 test images.}. The results are	reported in Table~\ref{tab:face_recovery}. It shows that for the image recovery with the top-half face images, the image generated with ACCA is of much better quality than that of MVAE and Bi-VCCA. 
		\begin{table}[t]
			\centering
			\renewcommand{\arraystretch}{1.3}
			\vspace{1.5mm}
			\hspace{-1mm}
			\caption{\small {FID (smaller is better) and sharpness (larger is better) scores for the image recovery on CelebA. The sharpness of the real images is 14.6722.}}
			\label{tab:face_recovery}
			\vspace{1.5mm}
			\scalebox{1}{
				\begin{tabular}{c|c|ccc}
					\toprule
					\multirow{2}{*}{\begin{tabular}[c]{@{}c@{}}\textbf{Input}\\ (halved image)\end{tabular}} & \multirow{2}{*}{\textbf{Methods}}& \multicolumn{3}{c}{\textbf{\;Evaluation metrics\qquad}} \\ \cline{3-5} & & FID & Sharpness  \\ \hline
					\multirow{3}{*}{\textbf{Top}} 
					& MVAE    & 61.3360 & 8.9645 \\ 
					& Bi-VCCA & 78.0752 & 7.0069  \\ 
					& ACCA    & \textbf{58.7983} & \textbf{11.9026}  \\ \hline
					\multirow{3}{*}{\textbf{Bottom}} 
					& MVAE    & \textbf{63.6921} & 8.5428  \\ 
					& Bi-VCCA & 84.7122 & 6.7574 \\
					& ACCA    & 68.1467 & \textbf{8.7249} \\ 
					\bottomrule
			\end{tabular}}
		\end{table}
	Bi-VCCA is the worst in terms of both the two metrics. For the experiment with the bottom-half face images, the FID score of our ACCA is slightly inferior than that of MVAE, however, the generated images of ACCA are still shaper. Comparing the results of these two experiments, we can see that the image recovery with the top-half image is better than the other, because it presents lower FID and higher image sharpness. This observation coincide with our qualitative evaluation shown in Fig.~\ref{fig:face_generation_result}, where the images generated with the bottom-half image (The left column), especially the top half generated images, is commonly blurrier then that generated with the top-half image (The right column). This phenomenon also agrees with our discovery in the hand written digit recovery task, where the input view with more information obtains worse results.
		\begin{figure}[t]
			\centering
			\hspace{-3mm}
			\includegraphics[width=0.9\textwidth]{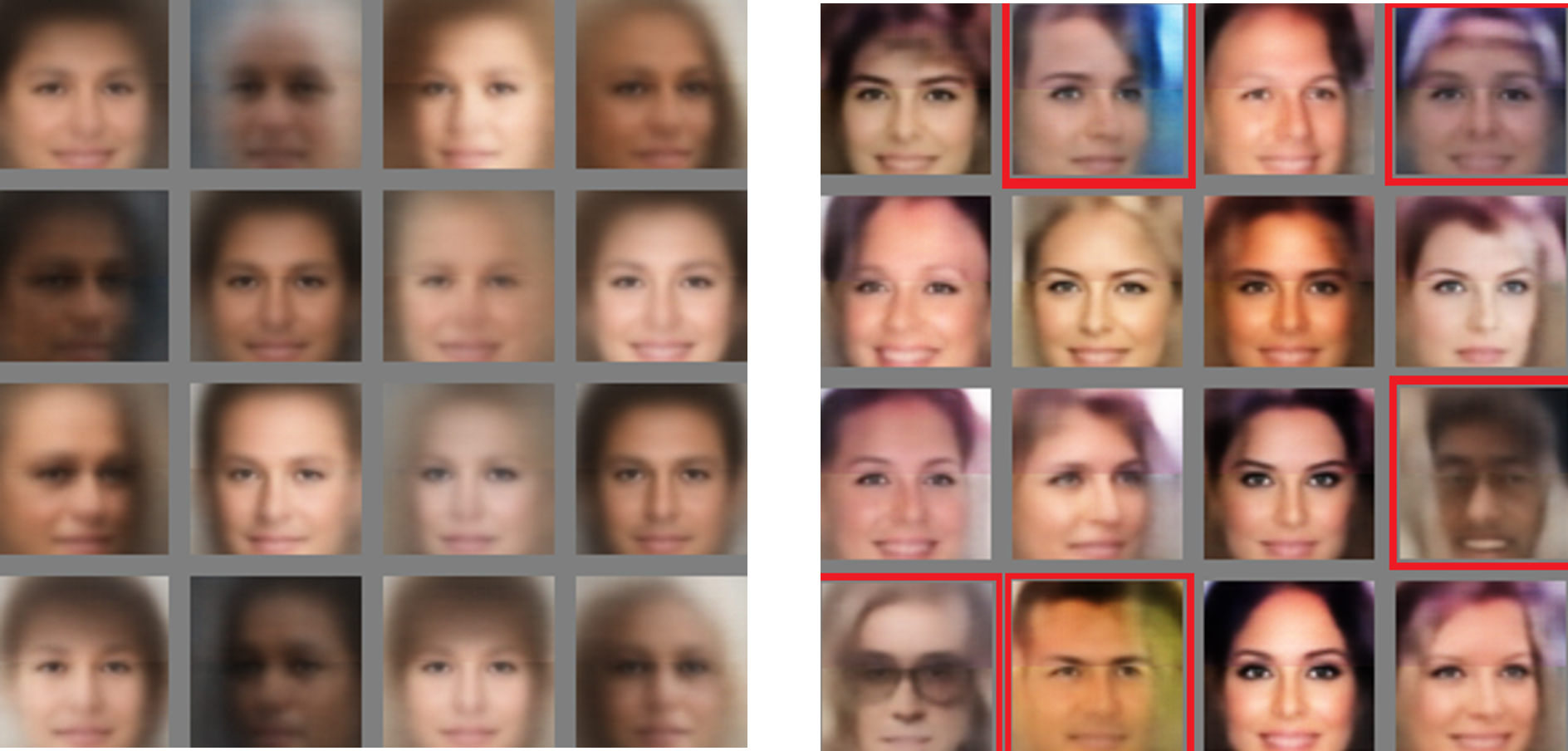}
			\caption{Comparison of Bi-VCCA [left] and ACCA [right] on unconditional generation. The images marked with red box present distinguishable details.  }\label{fig:face_un_generation_result}
			\vspace{-3mm}
		\end{figure}
%
	
\vspace{2mm}

\noindent\textbf{{Unconditional human face generation:}} {To illustrate how ACCA benefits the image generation quality, we further evaluate ``unconditional generation'' performance with the trained models on the CelebA dataset. Specifically, we randomly sample a batch of $z$ from the prior distribution $p(z)$ and adopt the two decoders to generate both views.}
{The results are presented in Fig.~\ref{fig:face_un_generation_result}. It is clear that the images generated with ACCA are much more realistic compared with Bi-VCCA, since facial boundaries of these images are more clear. It is also remarkable that the generates images with more details, due to the superior correspondence achieved between the input and the latent space. The images in red box present remarkable details, such as cap, hoodie, glasses and backgrounds.}  

	\subsubsection{Face alignment}
	We further evaluate the multi-view alignment performance of ACCA with face-alignment task ~\citep{kazemi2014one} on \textbf{\emph{CelebA}}~\citep{liu2015deep}. 
	We train ACCA with paired face and ground truth facial landmark annotations as input for the two views. Then, the better the multiple views are aligned, the better facial landmark prediction, or generation, results, can be achieved given the test face images.
	
	Since the landmarks annotations of the original CelebA dataset simply contains five landmark-locations, this dataset maybe insufficient to testify the performance achieved with these models that can handle more complicated applications~\citep{regmi2018cross}. Instead, we construct a more challenging dataset with 68 landmark locations as the face annotation. Specifically, we extract the annotations with the state-of-the-art facial landmark localization method Super-FAN~\citep{bulat2018super}, with the s$^{3}$fd face detector~\footnote{https://github.com/1adrianb/face-alignment}. We drop the figures whose faces cannot be detected and construct a dataset with 202,405 samples. Fig.~\ref{fig:CelebA_face_alignment_sample} presents several samples of our dataset. Details for the setting of the face alignment experiment is presented in Table~\ref{tab:dataset_generation}. 
		\begin{figure}[t]
			\centering
			\hspace{-3mm}
			\vspace{-2mm}
			\includegraphics[width=0.9\textwidth]{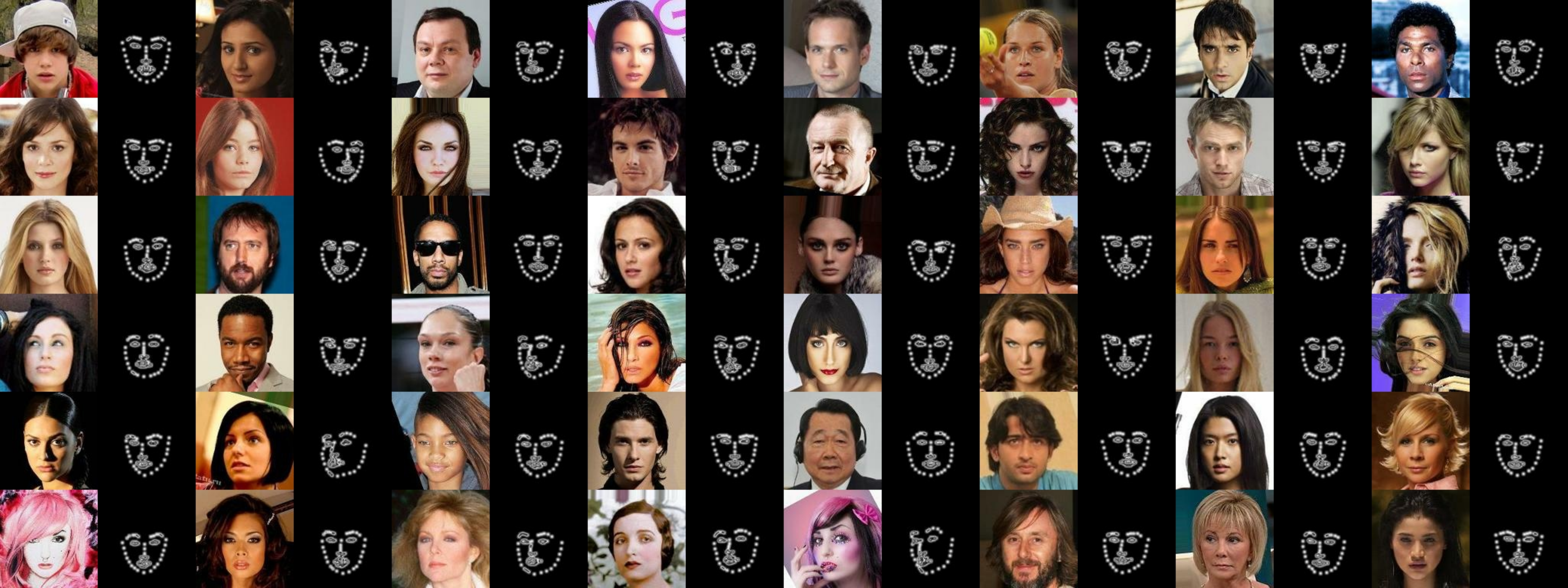}
			\caption{Sample images of CelebA for the face alignment experiment. }\label{fig:CelebA_face_alignment_sample}
			\vspace{-3mm}
		\end{figure} 

\vspace{-4mm}

		\textbf{\emph{Qualitative analysis}}: To verify the robustness of ACCA in achieving multi-view alignment, we adopt the complete data samples for training, while adopting partial or noisy images as input to evaluate the alignment performance of each model. Specifically, we randomly omit the input pixels with blocks of different sizes (50, 60, 70). Such setting simulates the real face alignment scenarios with occlusive faces. 
			\begin{figure}[t]
			\centering
			\hspace{-3mm}
			\vspace{-3mm}
			\includegraphics[width=\textwidth]{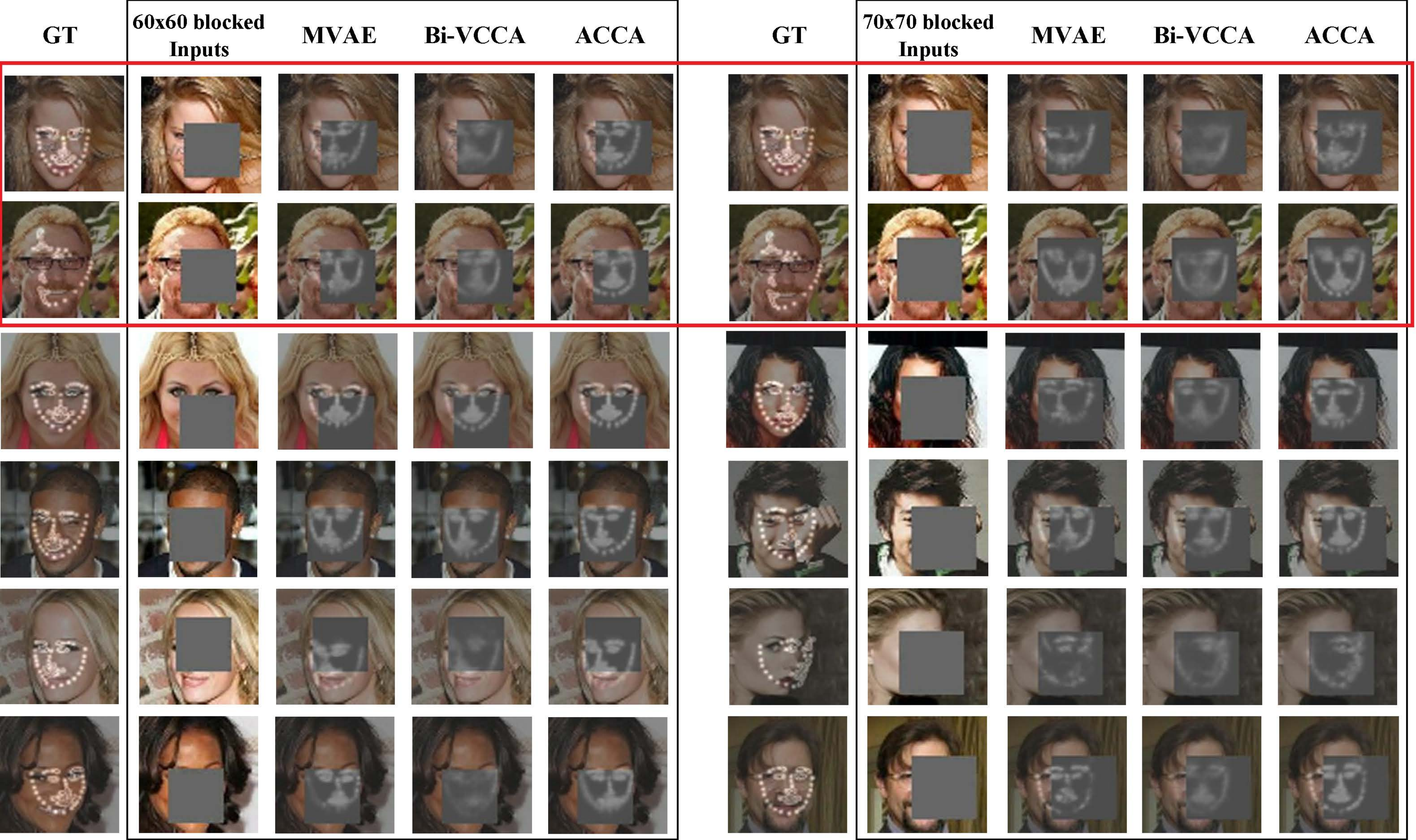}\caption{Performance of face alignment with different level of face occlusions.\textbf{Left:} the results with 60x60 blocked inputs; \textbf{Right:} the results with 70x70 blocked inputs. }\label{fig:face_alignment_results}
			\vspace{-2mm}
		\end{figure}
		
		Fig.~\ref{fig:face_alignment_results} demonstrates the face alignment results. It is clear that our proposed ACCA outperforms the baselines under both the two settings, with human interpretable and more clear facial landmark annotations. We can also observe that 
		\begin{itemize}
			\item[1).] Most of the generated results of MVAE are noisy and blurred under human perceptions, which indicates that MVAE is susceptible to noisy input. The problem is even more obvious with larger size of occlusions. As shown in the right column of the figure, most of the results of MVAE are not recognizable.
			\item[2).] The results of Bi-VCCA is commonly blurred than MVAE and our ACCA. However, Bi-VCCA is more robust with noisy input than MVAE since its results are more interpretable under the 70$\times$70 blocked setting. This verified that latent distribution matching constraint benefits the robustness of the multi-view alignment. 
			\item[3).] Our proposed ACCA achieves clear and human interpretable facial landmark annotations under both the two settings. This indicates that the multi-view alignment achieved with ACCA is the most robust among these three models. This verifies that the consistent encoding achieved in ACCA contribute to better and more robust alignment of the multiple views. 
			\vspace{-2mm}
		\end{itemize}
	 \noindent \textbf{\emph{Quantitative evidence}}: \:\: We further analyse the results with two standard metrics for image alignment, Peak Signal-to-Noise Ratio (\textbf{\emph{PSNR}})~\citep{bulat2018super} and Structural Similarity (\textbf{\emph{SSIM}})~\citep{Zhang_2018_ECCV}. Table~\ref{tab:face-alignment} shows that our ACCA is superior than the other two models with respect to both the two criteria. 
	 \begin{table}[t]
	 	\centering
	 	\vspace{-1mm}
	 	\hspace{-1mm}
	 	\renewcommand{\arraystretch}{1.2}
	 	\caption{\small PSNR (smaller is better) and SSIM (the larger is better) of face-alignment with random occlusions of different size. The best results are in bold. }
	 	\label{tab:face-alignment}
	 	\vspace{1.5mm}
	 	\scalebox{0.95}{
	 		\begin{tabular}{c|c|ccc}
	 			\toprule
	 			\multirow{2}{*}{\begin{tabular}[c]{@{}c@{}}\textbf{Evaluation metrics}\\ \end{tabular}} & \multirow{2}{*}{\textbf{Methods}}& \multicolumn{3}{c}{\textbf{Inputs} \ \ (occluded face images)} \\ \cline{3-5} 
	 			& & 50\,x\,50  & 60\,x\,60  & 70\,x\,70  \\ \hline
	 			\multirow{3}{*}{\textbf{PSNR}} 
	 			& MVAE    & 63.0074 & 62.5455 & 62.1448 \\ 
	 			& Bi-VCCA & 63.0289 & 62.6468 & 62.3351 \\
	 			& ACCA    & \textbf{62.2924} & \textbf{62.4175} & \textbf{62.0975} \\ \hline
	 			\multirow{3}{*}{\textbf{SSIM}} 
	 			& MVAE	  & 0.9982  & 0.9978 & 0.9975 \\ 
	 			& Bi-VCCA & 0.9981  & 0.9979 & 0.9976 \\ 
	 			& ACCA    & \textbf{0.9984}  & \textbf{0.9979} & \textbf{0.9981} \\ 
	 			\bottomrule
	 	\end{tabular}}
	 \end{table}

	\subsubsection{Cross-view generation for high-dimensional data}\label{sec:exp_clustering}
	
	{To evaluate the capacity of our ACCA for cross-view generation, we further validate its performance with high-resolution image inputs. We adopt the Google Maps dataset (Maps)~\citep{isola2017image} here, which is one of the benchmark datasets for cross-view synthesis applications~\citep{regmi2018cross}. To assure the quality of generated image, we equipped skip-connection for the autoencoder structure, i.e. UNET~\citep{ronneberger2015u}, in each method. We adopt the least square GANs~\citep{mao2017least} as the marginal matching constraint Eq.~\ref{eq:ACCA_three_appro}) for ACCA. The results are presented in Fig.~\ref{fig:cca_maps_results}. It is clear that our ACCA outperforms the baselines regarding the generated image quality.}
	\begin{figure}[t]
		\centering
		\includegraphics[width=0.98\textwidth]{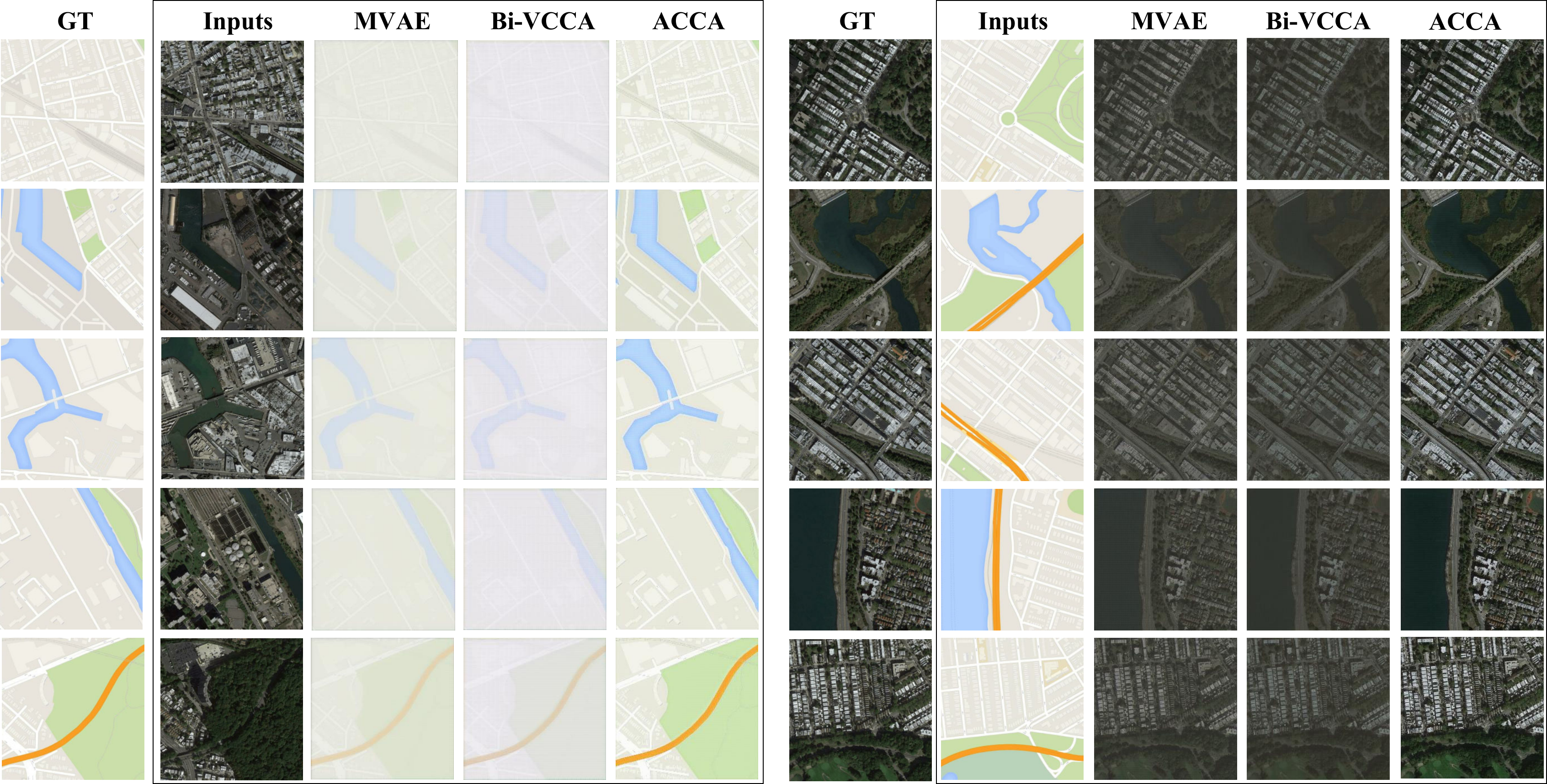}
		\vspace{-2mm}
		\caption{Comparison of cross-view generation results on the Maps dataset.}\label{fig:cca_maps_results}
	\end{figure}
\begin{figure}[hbtp]
	\hspace{-3mm}
	\centering
	\includegraphics[width=0.75\textwidth]{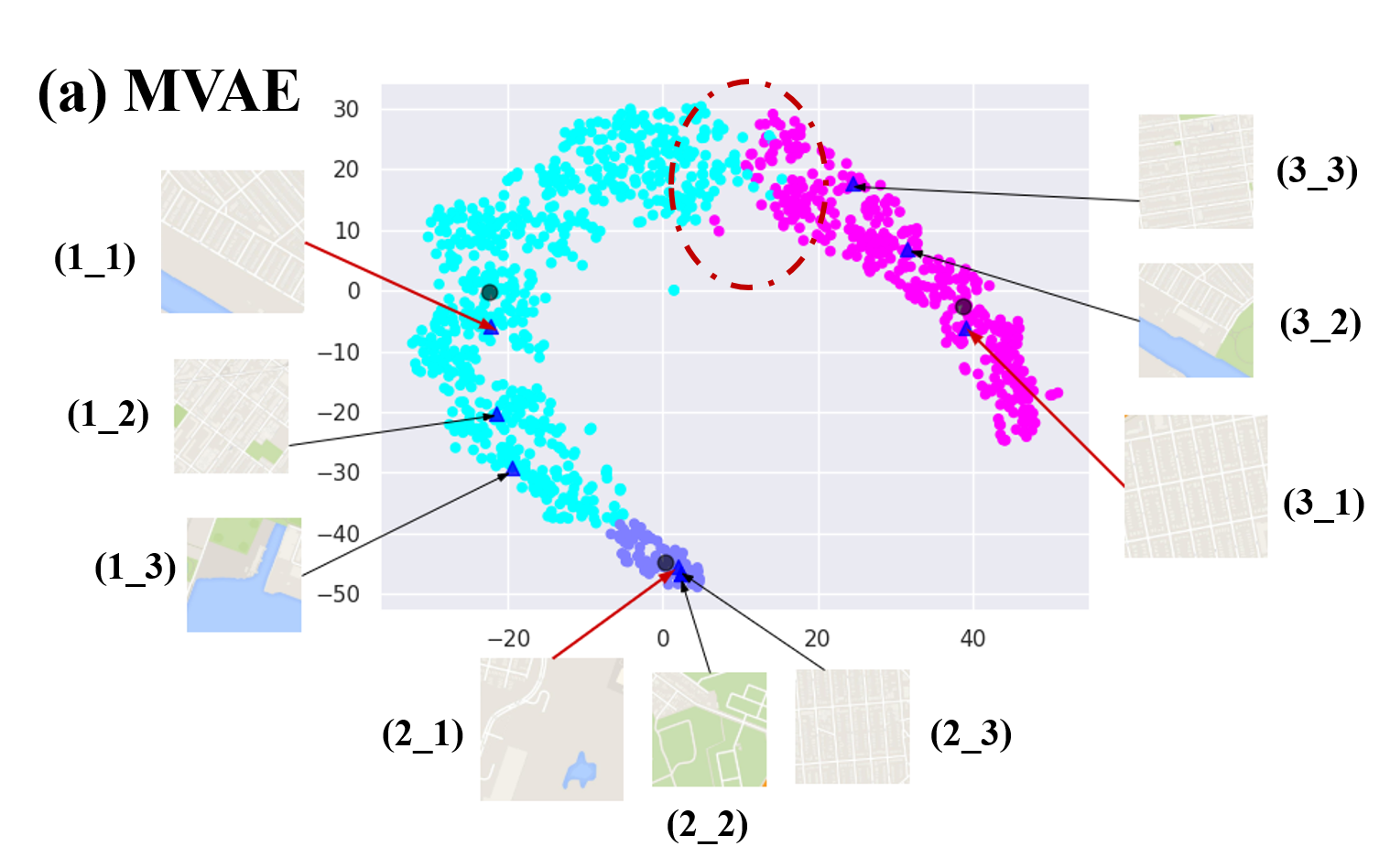}
	\vspace{0.5mm}
	\includegraphics[width=0.75\textwidth]{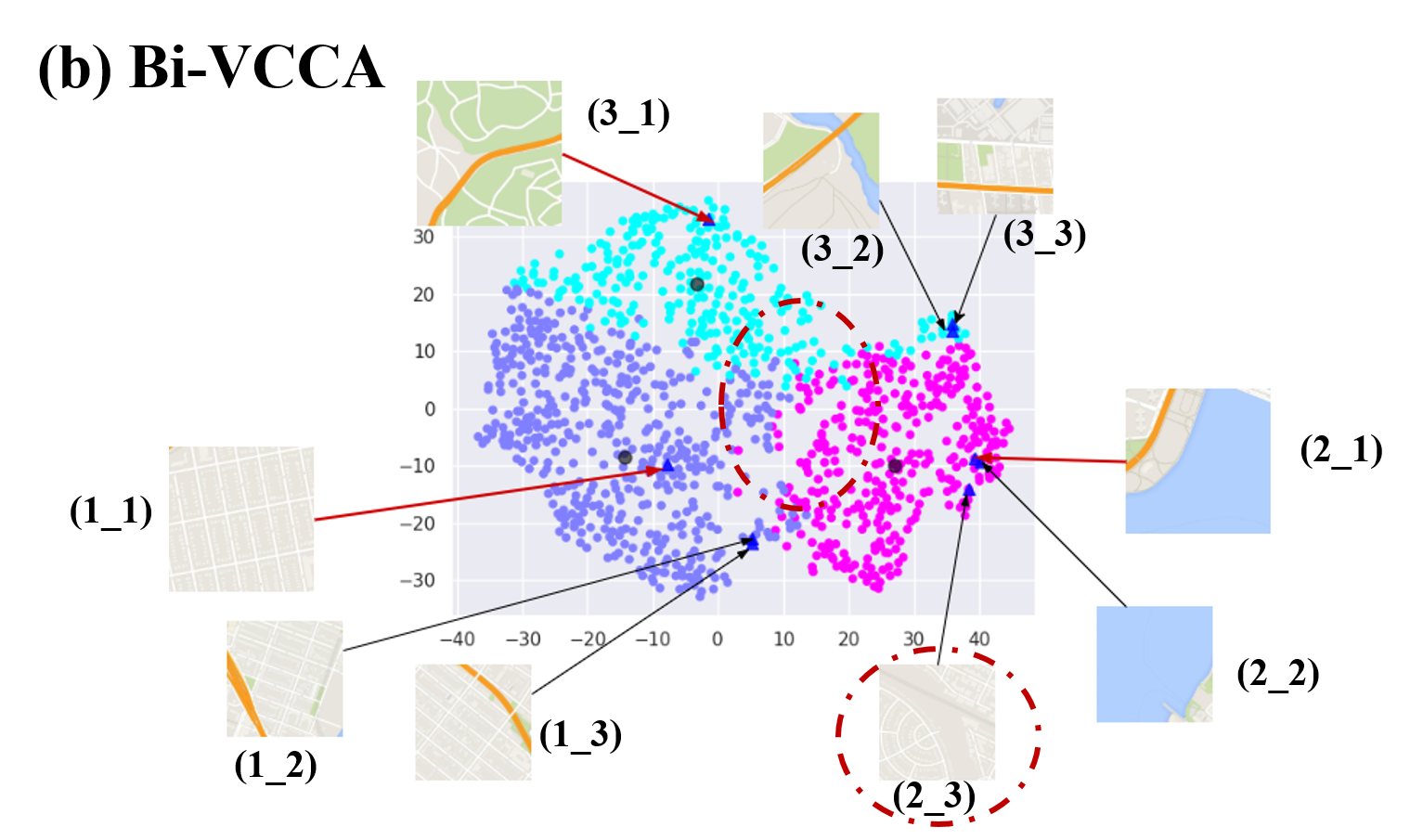}
	\vspace{0.5mm}
	\includegraphics[width=0.75\textwidth]{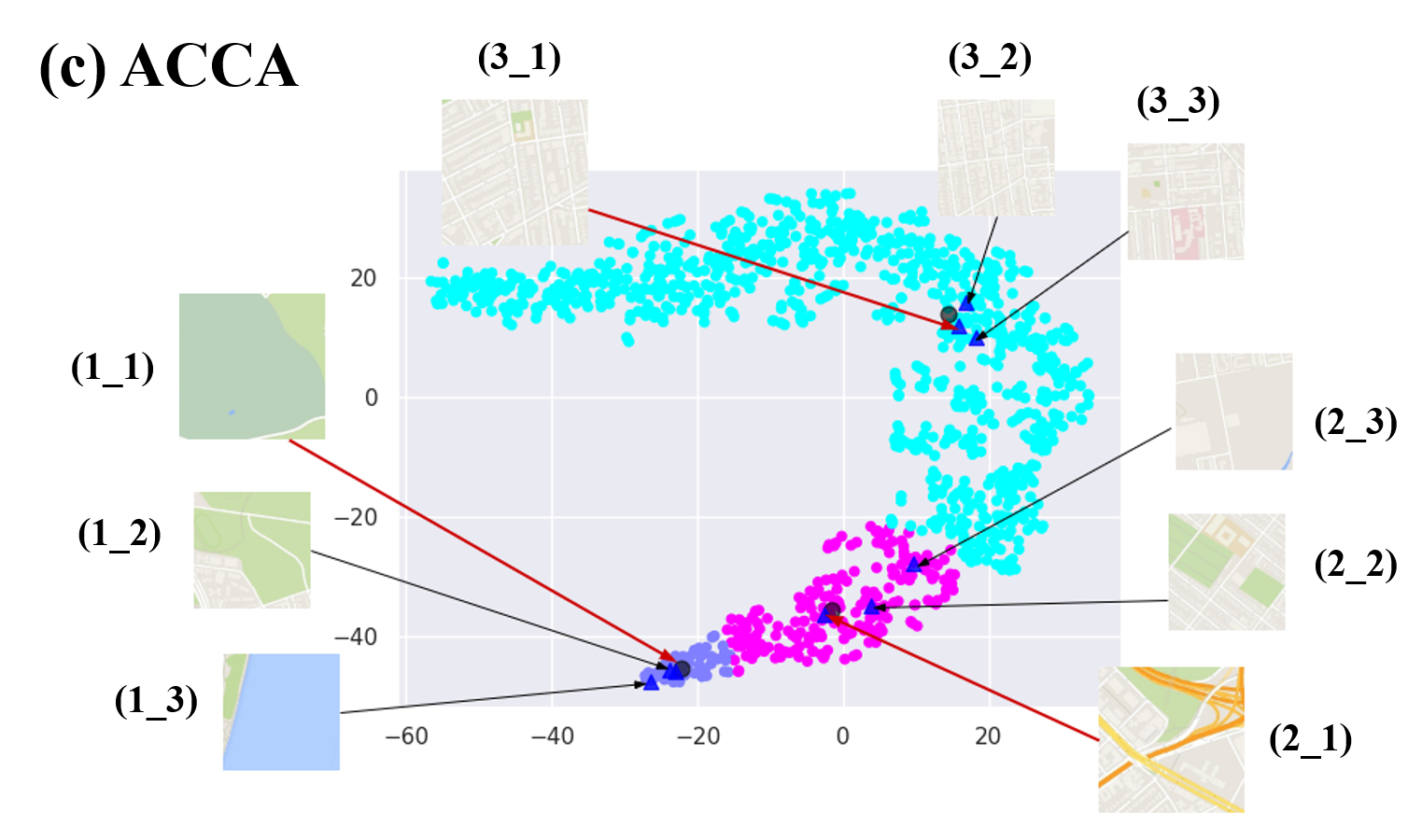}
	\vspace{-2mm}
	\caption{{Comparison of K-means clustering results on the embeddings of the Maps dataset. The centroids are marked with grey spots. The blue triangles represent the Top-3 data points nearest to each centroid, with the order represented in the annotations. For example, $1\_1$ is the data point nearest to the centroid within the first cluster.}}\label{fig:tsne_maps}
	\vspace{-3mm}
\end{figure}
\begin{figure}[t]
	\centering
	\includegraphics[width=0.96\textwidth]{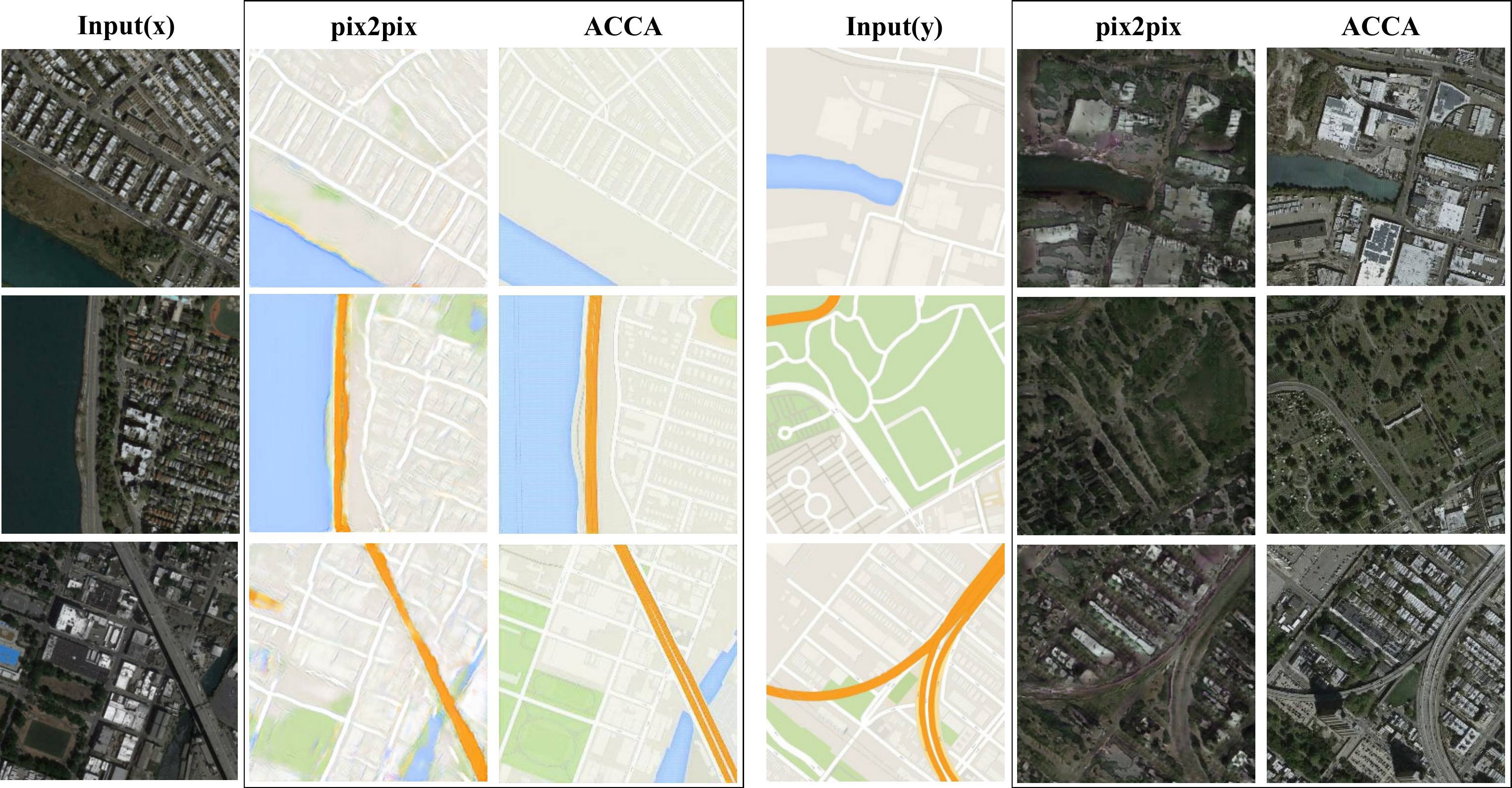} 
	\vspace{-2mm}
	\caption{Comparison of pix2pix and ACCA on the Maps dataset. }\label{fig:pix2pix_maps_results}
	\vspace{-4mm}
\end{figure}

{We also analysed the quality of obtained embeddings with t-SNE visualization. Specifically, we cluster the embeddings of the aerial photo [View$X$] with k-means{\small~\citep{macqueen1967some}}\mbox{(n\_clusters=3)} and do t-SNE visualization to analyse if the results present human interpretable properties. We mark the centroids of each cluster and prohibit the Top-3 data samples that are nearest to each centroid. The results are presented in Fig.~\ref{fig:tsne_maps}. The comparison is analysed in two aspects:}
\begin{itemize}
	\item[1).] {The clusters of our ACCA are compact and present clear boundaries, while that of MVAE and BI-VCCA both show overlap between the clusters (marked with circles of red dash lines.) The comparison is quite obvious with regards to Bi-VCCA. This indicates that the embedding of our ACCA preserves more discriminative information compared with Bi-VCCA.}
	\vspace{-2mm}
	\item[2).] {The clustering results of our ACCA represents human interpretable properties. According to Fig.~\ref{fig:tsne_maps}.(c), the three clusters presents distinct properties: among the test images, a large proportion of them are blocks (points in blue color), and small proportion are seas or vegetation (points in purple color), the rest are hybrid zones that contain highways, railways, etc. (points in red color). This discovery coincides with the data statistics of the original dataset. The clustering results of MVAE embeddings are not interpretable compared with our ACCA, since the samples nearest to the centroids are not distinguishable and there is no interpretable patterns presented for the Top-3 data samples for each cluster. The clusters centroids obtained with BI-VCCA is interpretable to some extent, since the data samples nearest to the centroids present unique properties. However, the clustering result of the data sample ``(2-3)'' circled with red is not understandable.}
\end{itemize}

{Consequently, on the Google Maps dataset, our ACCA outperforms the baselines regarding the alignment of the multiple views. The obtained latent embeddings are also informative about the multi-view data.}

{To collaboratively support the superior alignment and generation performance of ACCA, we further compare the result with pix2pix --- the state-of-the-art GAN-based cross-view generation baseline. The comparison is shown in Fig.~\ref{fig:pix2pix_maps_results}.We can see that the image quality of our ACCA is comparable and even better than that of pix2pix. This indicates that our ACCA posses good alignment and generation ability. 
}

\subsection{{Alignment and discriminative property of the representation}}\label{sec:exp_classification}

{Multi-view representation learning is an important scenario of CCA. In this section, we conduct classification tasks to verify that the alignment achieved in ACCA does not greatly influence the discriminative property of the learned representations. }

{We follow the setting in Table~\ref{tab:dataset} and perform classification on the three labelled datasets, MNIST\_LR, MNIST\_noisy and XRMB. We train linear SVM classifiers with the concatenation of obtained embeddings, and then evaluate its accuracy on the projected test set, i.e. $[Z_{X_{Te}}, Z_{Y_{Te}}]$. For iteratively optimized nonlinear CCA models, we selected the embeddings obtained from the last {\small$5$} epochs for evaluation. We compare ACCA with Gaussian prior, namely ACCA(G), here for a fair comparison. PCCA is not evaluated regarding classification since it should be comparable with the linear CCA.}

{Table~\ref{tab:classification} presents the classification results. It is obvious that our ACCA achieves comparable and even better classification performance among the CCA variants with the generative mechanism. The results of ACCA excels Bi-VCCA in all the settings and is comparable with MVAE in most of the settings. This reveals that our ACCA preserves considerable discriminative property of the embeddings while achieving superior alignment of the multiple views. Our ACCA is inferior to DCCA regarding the classification tasks. This is because, instead of targeting alignment for discriminative representation learning as in DCCA, our model focuses on reconstruction for data generation. For MNIST-LR dataset, ACCA outperforms DCCA to a large extent. This indicates that reconstruction can benefit discriminative representation learning in certain scenarios. The finding also coincides with the outstanding performance of MVAE, here.} 
 
 	\begin{table}[t]
 	\centering
 	\renewcommand{\arraystretch}{1.4}
 	\caption{{The classification accuracy and standard derivation (both in \%) with the obtained latent embeddings. The best results are in bold.}}
 	\hspace{-1mm}
 	\label{tab:classification}
 	\scalebox{0.95}{
 		\begin{tabular}{c|cc|ccc}
 			\toprule
 			\textbf{Datasets} & \textbf{CCA} & \textbf{DCCA}& \textbf{MVAE} & \textbf{Bi-VCCA} & \textbf{ACCA (G)} \\ \hline
 			MNIST\_LR   & 50.65 & 73.67$\pm$0.15 & 84.44$\pm$0.76 & 74.32$\pm$0.19  & \emph{\textbf{85.81$\pm$0.71}} \\
			MNIST\_Noisy& 75.48 & \textbf{91.60$\pm$0.36} & 90.78$\pm$1.12 & 85.81$\pm$0.44 & 86.93$\pm$1.46\\
			XRMB        & 32.04 & \textbf{62.14$\pm$0.52} & 58.57$\pm$0.30 & 56.58$\pm$0.35 &\emph{60.37$\pm$0.40}\\
 			\bottomrule
 	\end{tabular}}
 	\vspace{-1mm}
 \end{table}

\subsection{{ACCA vs CCA variants with view-specific information}}\label{sec:bivcca_private}

{In this section, we compare ACCA with CCA variants that {additionally} exploits view-specific information, to further demonstrate its alignment capacity. This is also a preliminary study on the influence of private information to multi-view alignment and generation. We choose Bi-VCCA-private as a representative baseline to compare here.} 
\begin{itemize}
	\item {\textbf{BI-VCCA-private}~\citep{wang2016deep}: An extension of~\emph{Bi-VCCA} \mbox{that additionally} extracts view-specific (\emph{private}) variables for each view. (Figure 2 in the paper)}
\end{itemize}

{
We evaluate the alignment in BI-VCCA-private regards to both correlation analysis and conditional/unconditional data generation. 
We compare our model with simple Gaussian prior, i.e. ACCA (G), here. We adopt the same settings, i.e. network settings and evaluation metrics, as previous experiments for consistency. For BI-VCCA-private, the dimension of private variables are set as $d_{H_{x}} = d_{H{y}} = 30$ for all the datasets. The network of the private encoders are set as the same as its principle encoders in Table~\ref{tab:dataset}.}

{{The results of correlation analysis are} presented in Table~\ref{tab:ACCA_vs_private}. It is clear that our ACCA excels Bi-VCCA-private and Bi-VCCA, in all the settings. This indicates that it is the KL-divergence constraint, i.e. $D_{KL}(q(z|*)\parallel p(z))$, that mainly hinder these models from {achieving instance-level multi-view alignment} (Figure 1.(b)). Our ACCA overcomes this limitation with the marginalized matching constraint, i.e. {$D_{JS}(\int q(z|*)p(*) d{*} \parallel p(z))$}, and thus preserves better correspondence for paired inputs. The results of Bi-VCCA-private are slightly better than that of Bi-VCCA{, indicating that the private variables can help to enhance multi-view alignment to some extent.}}

{The comparison regarding cross-view generation and unconditional data generation are presented in Figure~\ref{fig:face_generation_result_VCCA_private} and Figure~\ref{fig:bicca_private_face_generation_result}, respectively.
The results show that our ACCA outperforms the others in terms of both image sharpness and recognisable object details. In cross-view generation, the images generated with our ACCA are much clear and sharper than Bi-VCCA-private. In addition, although Bi-VCCA-private generates faces with more details compared with Bi-VCCA, e.g. beards and glasses (Figure~\ref{fig:bicca_private_face_generation_result}), these details are not as clear and recognisable as that of our ACCA. These generation results coincide with our finding in correlation analysis--- Bi-VCCA-private achieves inferior multi-view alignment compared with ACCA. The inferior alignment in Bi-VCCA-private consequently downgrades its performance in cross-view data generation.}

{Furthermore, it is interesting to notice that, for Bi-VCCA-private, the image quality generated with different views are of slight difference, compared with the baselines that only the extract shared information~(Figure~\ref{fig:face_generation_result}). {This indicates that incorporating view-specific variables also contributes to a balanced cross-view generation capacity from different views, when the multiple input views contain an imbalanced amount of information. (In Section~\ref{sec:image_recovery}, we find that the imbalance of information between the two input views can influence the image generation quality)}}
	 \begin{table}[t]
	\centering
	\hspace{-1mm}
	\renewcommand{\arraystretch}{1.2}
	\caption{\small {The dependency (higher is better) of latent embeddings. The best are in bold.} }
	\label{tab:ACCA_vs_private}
	\vspace{1.5mm}
	\scalebox{0.85}{
		\begin{tabular}{c|c|ccc}
			\toprule
			\multirow{2}{*}{\begin{tabular}[c]{@{}c@{}}\textbf{ Metrics}\\ \end{tabular}} & \multirow{2}{*}{\textbf{Methods}}& \multicolumn{3}{c}{\textbf{Datasets}} \\ \cline{3-5} 
			& & MNIST\_LR  & MNIST\_Noisy  & XRMB  \\ \hline
			\multirow{3}{*}{{\begin{tabular}[c]{@{}c@{}}\\ \textbf{nHSIC}\\(linear kernel)\end{tabular}}}
			& \textbf{BI-VCCA-private}  & 0.2818 & 0.2235& 0.1227 \\ \cline{2-5} 
			& \textbf{Bi-VCCA}     & 0.4612& 0.1912 & 0.1046\\ 
			& \textbf{ACCA (G) (\emph{ours})}     & \textbf{0.5423} & \textbf{0.3285} & \textbf{0.2903} \\ 
			\hline
			\multirow{3}{*}{{\begin{tabular}[c]{@{}c@{}} \\ \textbf{nHSIC}\\(RBF kernel)\end{tabular}}}
			& \textbf{BI-VCCA-private} 	& 0.2853  & 0.2386  & 0.0893 \\ \cline{2-5}
			& \textbf{Bi-VCCA}     & 0.3804& 0.2076 & 0.0846\\ 
			& \textbf{ACCA (G) (\emph{ours})}    & \textbf{0.6318 } & \textbf{0.3099} & \textbf{0.2502} \\ 
			\bottomrule
	\end{tabular}}
\end{table}
\begin{figure}[t]
	\centering
	\hspace{-4mm}
	\includegraphics[width=0.9 \textwidth]{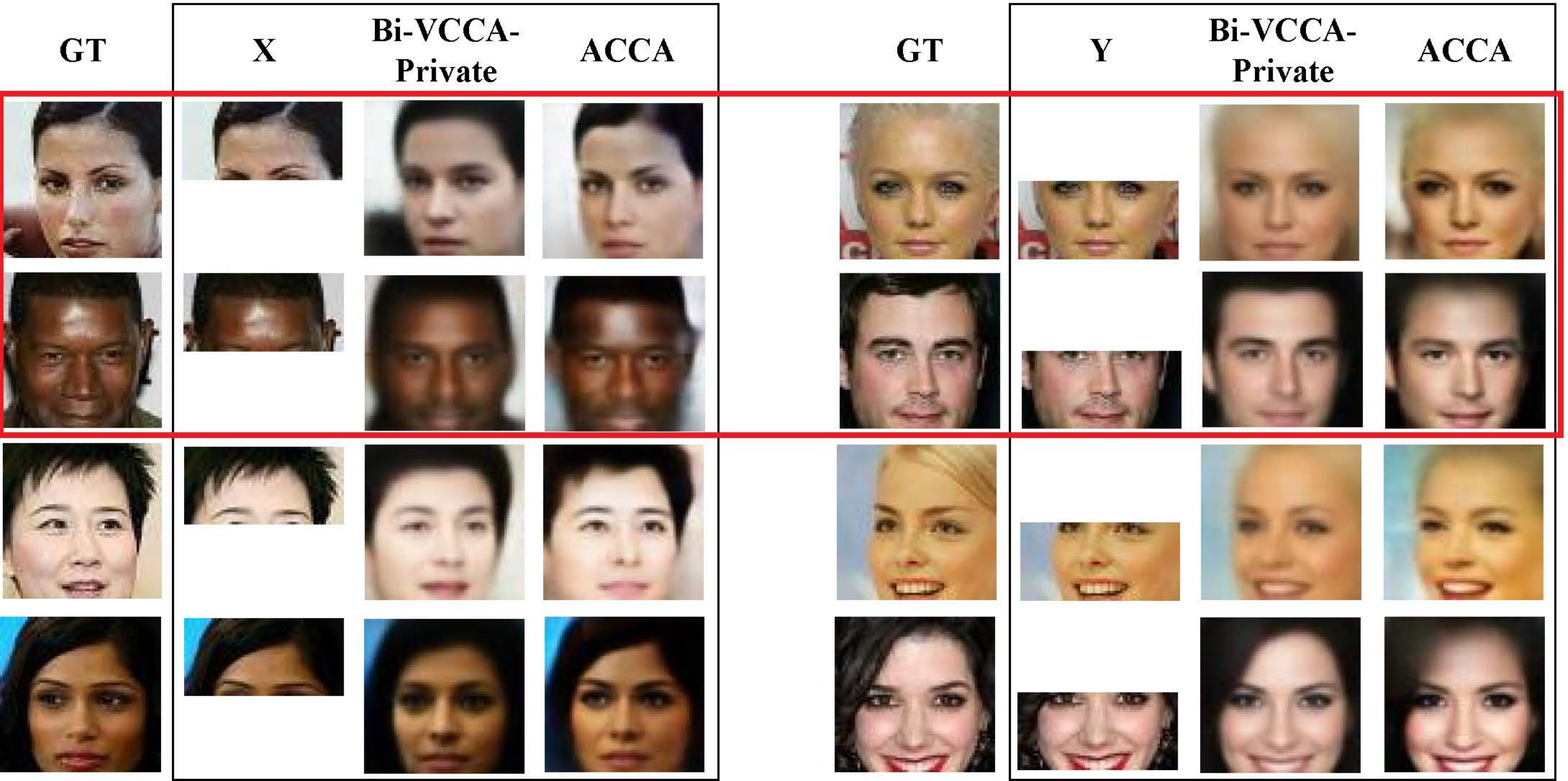}
	\vspace{-2mm}
	\caption{{Comparison of Bi-VCCA-private and ACCA regarding the face recovery. The results of Bi-VCCA-private is commonly blurred than our ACCA.} }\label{fig:face_generation_result_VCCA_private}
	\vspace{-1mm}
\end{figure}
\begin{figure}[t]
	\centering
	\hspace{-2mm}
	\includegraphics[width= 0.98
	 \textwidth]{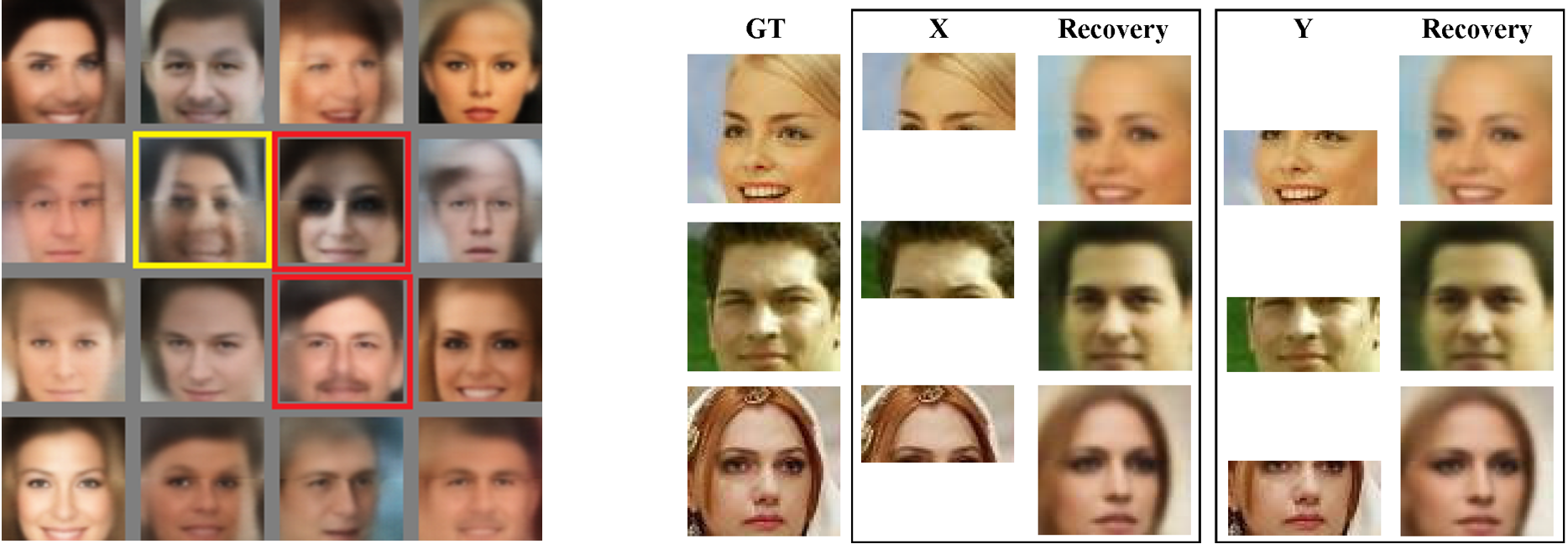}
	\vspace{-2mm}
	\caption{{Analysis on the face-recovery results of Bi-VCCA-private. \textbf{[Left]:} The unconditional generation results of Bi-VCCA-private. It is inferior to  our ACCA, although superior to Bi-VCCA (Figure~\ref{fig:face_un_generation_result}) 
	\textbf{[Right]:} For Bi-VCCA-private, the image quality generated with different views are of slight difference. This indicates that incorporating private variables contributes to a balanced capacity for data generation from each view.} }\label{fig:bicca_private_face_generation_result}
\end{figure}

	\section{Conclusion} \label{sec:conclusions}
	
	In this paper, we present a systematic analysis of instance-level multi-view alignment with CCA. Based on the marginalization principle of Bayesian inference, We propose to study multi-view alignment via consistent latent encoding and present ACCA which facilitate superior alignment of the multiple views that benefits the performance of various of multi-view analysis and cross-view analysis tasks. Matching multiple encodings, ACCA can also be adopted to other tasks, such as image captioning and translation. Furthermore, owing to the flexible architecture design of our ACCA, it can be easily extended to multi-view task of $n$ views, with ($n+1$) encoders and ($n$) decoders. 
	
    {In this work, we mainly exploit our ACCA with pre-defined priors. For future work, we will explore more powerful inference techniques for ACCA to further boost its alignment performance. For example, normalizing flows~\citep{rezende2015variational} is a data-driven method, that provides an efficient tool to learn a data-dependent prior for complex datasets. It can be employed into ACCA to boost multi-view alignment by providing a more expressive latent space, i.e. with a complex data-dependent prior, and better preservation of instance-level correspondence,  i.e. with invertible mappings.}
	
	\vspace{0.3mm}
	
	Our analysis based on the CMI and the consistent encoding also provides insights for a flexible design of other CCA models. In the future, we will conduct more in-depth analysis on multi-view alignment with CMI and propose other variants of CCA with other alignment criteria, e.g. MMD distance, Wasserstein distances\citep{DBLP:journals/corr/ArjovskyCB17}. It is also interesting to notice that input with different-level of details can influence the result of cross-view generation. This research direction is also worth further study.  

\subsection*{Acknowledgments}
The work was supported  in part by the Australian Research Council grants DP180100106, DP200101328 and China Scholarship Council (No.201706330075)
			\begin{table}[t]
		\centering
		\large  
		\caption{Details of the cross-view generation datasets.} \label{tab:dataset_generation}
		\vspace{1mm}
		\setlength{\tabcolsep}{1.15mm}{
			\scalebox{0.6}{%
				\renewcommand{\arraystretch}{1.2}
				\begin{tabular}{|c|c|c|c|c|c|c|}
					\hline
					Dataset & Statistics & \tabincell{c}{\tabincell{c}{Dimension\\ of ${z}$}} & \multicolumn{2}{c|}{\tabincell{c}{Architecture
							\\(Conv all with batch normalization before LReLU)}} &Parameters\\ \hline
					\tabincell{c}{CelebA\\\citep{liu2015deep} } & \tabincell{c}{ \textbf{Image Resolution:} \\
						64 $\times$ 64 \\ \\ \textbf{Image recovery} \\ \# Tr= 201,599 \\ \# Te= 1,000 \\\\ \textbf{Face alignment} \\ \# Tr= 201,599 \\ \# Te= 1,000 \\ 
						
					} & d = 100 &\multicolumn{1}{c|}{\tabincell{c}{\textbf{Encoders}:\\Conv: 64$\times$5$\times$5 (stride 2), \\ Conv: 128$\times$5$\times$5 (stride 2),\\Conv: 256$\times$5$\times$5 (stride 2),\\Conv: 512$\times$5$\times$5 (stride 2);\\dense: 100.\\ \textbf{Decoders (Image recovery)}:\\dense: 8192, relu;\\deConv: 256$\times$5$\times$5 (stride 2),\\deConv: 128$\times$5$\times$5 (stride 2),\\deConv: 64$\times$5$\times$5 (stride 2),\\deConv: 3$\times$2$\times$5,  (stride1$\times$2);\\Tanh.\\ \textbf{Decoders(Face alignment)}:\\dense: 8192, relu; \\deConv: 256$\times$5$\times$5 (stride 2),\\deConv: 128$\times$5$\times$5 (stride 2),\\deConv: 64$\times$5$\times$5 (stride 2), \\deConv: 3$\times$2$\times$5 (stride 2);\\Tanh.}}&\tabincell{c}{\textbf{Discriminator: $\hat{D}$}:\\dense: 128$\rightarrow$64$\rightarrow$1,\\
						sigmoid.\\}& \tabincell{c}{Epoch = 10;\\ Batchsize = 64;\\ lr= 0.0002;\\ Beta1 = 0.05;}\\   \hline
					\tabincell{c}{Google Maps dataset \\ (Maps) \\ \citep{isola2017image}} & \tabincell{c}{ \textbf{Image Resolution:} \\
						256 $\times$ 256 \\ \\ \textbf{Cross-view generation} \\ \# Tr= 1,096 \\ \# Te= 1,098 \\
					} & d = 100 &\multicolumn{1}{c|}{\tabincell{c}{\textbf{Encoders}:\\Conv: 64$\times$5$\times$5 (stride 2),\\Conv: 128$\times$5$\times$5 (stride 2),\\Conv: 256$\times$5$\times$5 (stride 2), \\Conv: 256$\times$5$\times$5 (stride 2),\\Conv: 512$\times$5$\times$5 (stride 2);\\dense: 100.\\ \textbf{Decoders (with skip-connection)}:\\dense: 32768,relu;\\deConv: 256$\times$5$\times$5 (stride 2),\\deConv: 256$\times$5$\times$5 (stride 2),\\deConv: 128$\times$5$\times$5 (stride 2),\\deConv: 64$\times$5$\times$5 (stride 2),\\deConv: 3$\times$2$\times$5,  (stride 2);\\Tanh. }}&\tabincell{c}{\textbf{Discriminator: $\hat{D}$}:\\dense: 128$\rightarrow$64$\rightarrow$1,\\
						tanh.\\}& \tabincell{c}{Epoch = 15;\\ Batchsize = 16;\\ lr= 0.0002;\\ Beta1 = 0.5;}\\
					\hline
		\end{tabular}}}
		\vspace{-2mm}
	\end{table}

		\bibliographystyle{APA}
		\bibliography{ICCA}   
		%
		%
		%
		%
	\end{document}